\documentclass{article}

\PassOptionsToPackage{numbers, compress}{natbib}
\usepackage[preprint]{neurips_2026}
\usepackage{graphicx}
\usepackage[utf8]{inputenc}
\usepackage[T1]{fontenc}
\usepackage{hyperref}
\usepackage{url}
\usepackage{booktabs}
\usepackage{amsfonts}
\usepackage{nicefrac}
\usepackage{microtype}
\usepackage{multirow}
\usepackage{xcolor}
\usepackage{multirow}
\usepackage{xcolor}

\usepackage{multirow}
\usepackage[table]{xcolor}
\usepackage{array}
\usepackage{booktabs}
\usepackage{amsmath} 

\usepackage{array}
\usepackage[table]{xcolor}
\newcolumntype{V}{!{\vrule width 0.35pt}}
\usepackage{enumitem}
\newcommand{\scell}[2]{\shortstack[c]{#1 / #2}}
\newcommand{\sblank}[1]{\shortstack[c]{#1 / \,}}

\newcolumntype{V}{!{\color{gray!55}\vrule width 0.2pt}}

\title{Multi-Head Latent Control: A Unified Interface for LLM Agent Decision Making}

\author{%
  Amirhosein Ghasemabadi\textsuperscript{1,*},
  Ruichen Chen\textsuperscript{1},
  Bahador Rashidi\textsuperscript{2},
  Di Niu\textsuperscript{1}
  \\[2pt]
  \textsuperscript{1}ECE Department, University of Alberta
  \\
  \textsuperscript{2}Huawei Technologies Canada Co., Ltd.
  \\[2pt]
  \texttt{\{ghasemab, ruichen1, dniu\}@ualberta.ca}
  \\
  \texttt{rashidibahador92@gmail.com}
  \\[2pt]
  \textsuperscript{*}Corresponding author
}

\begin{document}

\maketitle

\begin{abstract}
Large language models are increasingly deployed as agents, but reliable agentic behavior requires more than next-token prediction. At inference time, it is preferred that an agent can decide whether to proceed with its current reasoning, defer to a stronger model, request additional information, invoke external tools, or abstain under the given setup. Existing approaches address these decisions through prompt-level routing, external orchestration, or task-specific fine-tuning, which primarily rely on input-side signals, and are often costly and difficult to maintain as model backbones evolve. We ask whether such control decisions can be inferred directly from a model’s latent generation process. We introduce \textbf{Multi-Head Latent Control}, a lightweight layer that reads hidden-state trajectories from a frozen LLM or VLM to produce deployment-time control signals. A \textbf{Capability Head} predicts whether the current model can solve the instance or should defer to a stronger collaborator, while a \textbf{Resolution Head} predicts appropriate resolution decision \textit{Clarification}, \textit{Tool Use}, \textit{Abstention}, or \textit{Direct Answering}. Both heads are trained only on latent traces from the same frozen LLM backbone, enabling post hoc adaptation without modifying the model. 
Across language and vision-language settings, Multi-Head Latent Control consistently improves the quality–cost tradeoff of multi-model systems, enabling early handoff from partial generations and more accurate intervention decisions. In routed execution (small + large model), it reduces large-model usage by up to 90.7\% on AndroidWorld and 27–53\% on average across benchmarks, while retaining most of large-model performance. Additionally, the learned control signals improve tool-use decision quality, yielding up to +158\% relative score gain and 65.5\% fewer missed-required tool calls.
\par\medskip
{\footnotesize
\noindent\textbf{Code:} \url{https://github.com/Amirhosein-gh98/Multi-Head-Latent-Control}
}
\end{abstract}

\section{Introduction}
Foundation models are increasingly capable, but reliable deployment requires more than strong next-token prediction. In long-horizon, multi-step, and tool-augmented agentic settings, an LLM must reason not only about what to generate, but also whether it can solve the current task or should defer to a stronger model. It must further determine whether additional information is required, whether external tools should be invoked, or whether the task is infeasible under the current setup. Errors in these decisions or overly conservative strategies such as always invoking tools or using a larger LLM for tasks are costly, leading to unnecessary computation, increased latency, avoidable external calls, and compounding downstream failures.

This challenge is particularly evident in emerging commercial agentic systems built on frontier models (e.g., Anthropic’s Claude agents), where token consumption remains high even for relatively simple tasks. This inefficiency suggests substantial untapped potential for reducing cost by enabling self-awareness in agentic LLMs allowing models to assess their own capability and act accordingly during inference.

Existing work addresses parts of this problem, but mostly through external mechanisms. Prompt- and category-level routing methods~\citep{chen2024frugalgpt,ong2025routellm,chen2024routerdc,dekoninck2025cascade} learn to assign queries to models or pipelines based on input-side signals such as prompt features or predicted task categories, improving coarse cost–quality tradeoffs but without assessing instance-level model adequacy.
Multi-agent collaboration and orchestration methods~\citep{du2024debate,hong2024metagpt,wang2025mixtureofagents,su2025toolorchestra} improve capability through interaction among models, tools, or roles, but often rely on heavier scaffolding or dedicated end-to-end training to make models better collaborators or orchestrators. As stronger LLMs and VLMs are released at an accelerating pace, repeatedly Fine-tuning each new backbone for such roles fails to keep pace with the evolution of foundation models. 

An orthogonal yet complementary line of work focuses on decoding-side methods, such as speculative decoding and reward-guided decoding strategies~\citep{leviathan2023speculative,shen2024collaborative,liao2025rewardguided}, which improve efficiency during generation by accelerating large-model inference once it is invoked. In contrast, our goal is to reduce how often large models are invoked by making deployment-time control decisions. Rather than replacing decoding-level optimization, we introduce a complementary layer: a lightweight intrinsic control interface that derives deployment-time decisions directly from the model’s latent states.
\begin{figure*}[t]
    \centering
    \includegraphics[width=\textwidth]{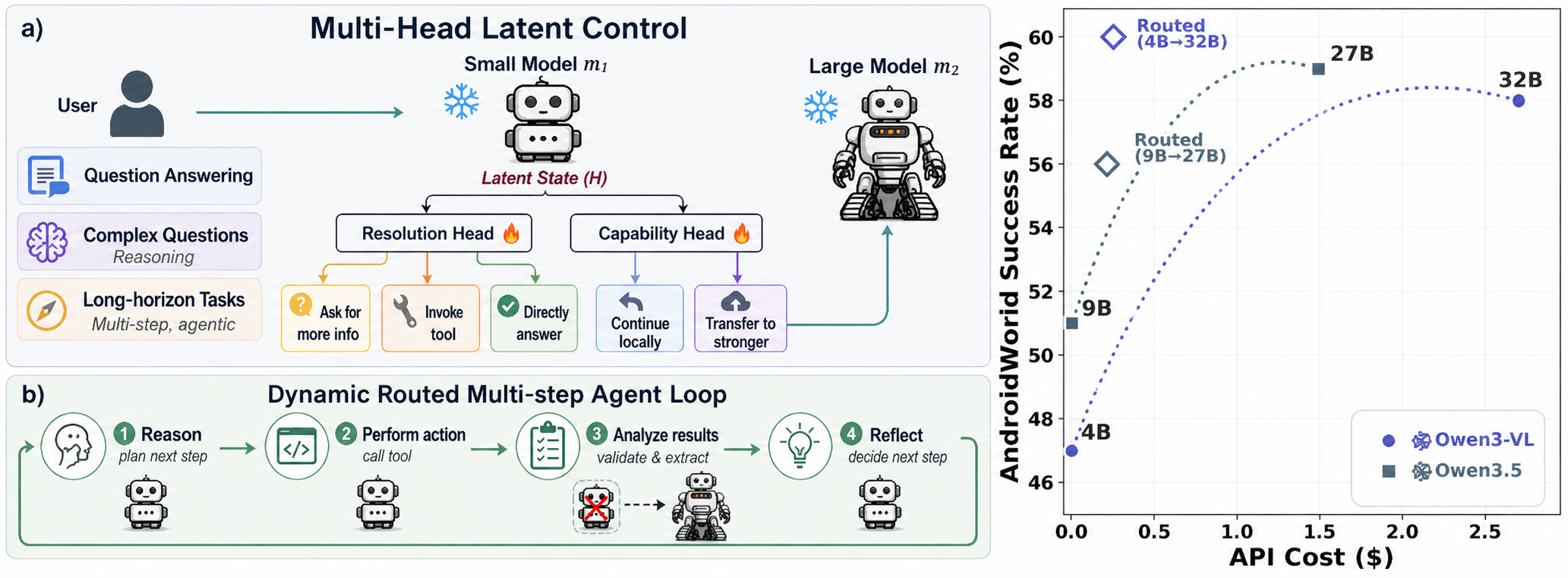}
    \caption{\textbf{Multi-head latent control as an intrinsic control interface for frozen foundation models.} \textbf{(a)} From the latent trajectory of a frozen primary model $m_1$, a \emph{Capability Head} decides whether to remain with $m_1$ or hand off to a stronger model $m_2$, while a \emph{Resolution Head} selects the appropriate resolution decision: request more information, invoke tools, abstain, or proceed directly. \textbf{(b)} Repeated use of these control signals enables dynamic routed execution in long-horizon, multi-step agent loops, allowing the system to choose the appropriate model at each step instead of relying on a large model throughout the entire trajectory. \textbf{(c)} On AndroidWorld, routed execution substantially improves the quality-cost tradeoff over single-model baselines by using the large model selectively.}
    \vspace{-1em}
    \label{fig:main}
\end{figure*}

In this paper, we show that frozen foundation models can expose a practical control interface for deployment-time decision-making without fine-tuning the backbone. We introduce \textit{Multi-Head Latent Control}, consisting of two lightweight heads: a \textbf{Capability Head} and a \textbf{Resolution Head}, which attach to a frozen language or vision-language model and emit actionable control signals during inference.
By reading hidden states from the model’s own generation process, the \textbf{Capability Head} predicts whether the current model should retain control or transfer the instance to a stronger fallback model. When the decision is to remain with the primary model, the \textbf{Resolution Head} determines whether the model should request additional information, call external tools, or abstain when the task cannot be answered under current setup. This introduces an effective form of self-awareness in routed agentic settings, improving the quality–cost tradeoff of inference and strengthening resolution decision-making, while remaining lightweight enough for rapid post hoc adaptation to new backbones. Our main contributions are:
\vspace{-0.5em}
\begin{itemize}[leftmargin=1.2em]

\item \textbf{A post hoc self-awareness layer for frozen LLMs.} 
We formulate deployment-time decision-making as a latent-state control problem, introducing a two-level self-awareness mechanism based on hidden-state trajectories. This enables a unified control interface for both model selection and intervention, without modifying or fine-tuning the backbone.
\vspace{-0.4em}

\item \textbf{Capability Head for model adequacy.} 
We introduce a lightweight {Capability Head} that reads hidden-state signals to predict whether the current model is adequate to solve a given instance, enabling reliable instance-level decisions on whether to retain control or escalate to a stronger model.
\vspace{-0.4em}

\item \textbf{Resolution Head for intervention decisions.}
conditioned on retaining control with the primary model, we introduce a Resolution Head that determines the appropriate intervention, including requesting additional information, invoking tools, abstaining under the current setup, or proceeding directly.
\vspace{-0.4em}

\item \textbf{Significant quality--cost gains with minimal adaptation.} 
Across model families and benchmarks, our method substantially reduces reliance on large models while preserving most of their performance. In routed small-large LLM agentic execution, it achieves up to \textbf{90.7\%} reduction in paid API cost on AndroidWorld and \textbf{27--53\%} average cost reduction across benchmarks. The \textbf{Resolution Head} further improves \textsc{When2Call} by up to \textbf{11.7} F1 points and \textbf{12.4} accuracy points, and enhances tool-use decisions on TriviaQA with up to \textbf{+158.9\%} relative score gain and \textbf{65.5\%} fewer missed-required tool calls.
\vspace{-0.8em}

\end{itemize}

\section{Related Work}
\vspace{-1em}
\paragraph{Model routers and adaptive selection.}
A growing line of work studies how to route queries across models or inference paths under quality, latency, and cost constraints. Methods such as \textit{FrugalGPT}~\citep{chen2024frugalgpt}, \textit{RouteLLM}~\citep{ong2025routellm}, \textit{RouterDC}~\citep{chen2024routerdc}, and cascade-style approaches~\citep{dekoninck2025cascade} learn routers or cascades from input-level signals, such as prompt characteristics, predicted task category, or other pre-generation cues. These methods improve coarse cost-quality tradeoffs, but they do not directly ask whether the current model is adequate for the specific instance at hand. Our setting is different: we study a deployment-time control signal read directly from the model's own generation process.
\vspace{-0.8em}

\paragraph{Multi-model collaboration and agentic systems.}
Another line of work improves performance by coordinating multiple models, tools, or specialized roles. Multi-agent debate~\citep{du2024debate}, role-based systems such as \textit{MetaGPT}~\citep{hong2024metagpt}, mixture-style systems~\citep{wang2025mixtureofagents}, and trained orchestration models~\citep{su2025toolorchestra} show that structured collaboration can improve reasoning and task performance. However, these approaches typically rely on external orchestration, fixed role assignments across models or agents, or dedicated training to make models better collaborators or orchestrators. Our approach instead asks whether a frozen model can be made into a better collaborator through a lightweight post hoc control layer.
\vspace{-0.8em}

\paragraph{Tool use and intervention decisions.}
A related literature studies when systems should call tools or switch to external resources. Work such as \textit{Toolformer}~\citep{schick2023toolformer}, \textit{Gorilla}~\citep{patil2024gorilla}, and \textit{ToolLLM}~\citep{qin2024toolllm} primarily improves tool-use behavior by training or adapting the model itself to invoke tools more accurately. Later work such as \textit{When2Call}~\citep{ross2025when2call}, \textit{MeCo}~\citep{li2025meco}, and related methods~\citep{xu2025efficienttool,xu2025reliabilityalignment} studies whether tool use is necessary and how to improve tool-calling decisions. Our setting is different: rather than training the backbone to become a better tool user, we attach a lightweight post hoc control layer to a frozen model. More broadly, our method goes beyond tool use alone by exposing a shared control interface that jointly governs transfer, clarification, tool use, abstention, and direct answering.
\vspace{-0.8em}

\paragraph{Decoding-time efficiency.}
Another adjacent line of work improves efficiency during generation. Speculative decoding and related methods accelerate inference by using a smaller draft model to speed up decoding by a larger target model~\citep{leviathan2023speculative,chen2023speculative,cai2024medusa,narasimhan2025faster,shen2024collaborative,liao2025rewardguided}. Their goal is to speed up large-model inference, not to reduce how often a large model is invoked. Therefore, they address a different and orthogonal problem and can naturally be combined with our method.
\vspace{-0.8em}

\paragraph{Reliability based on latent signals.}
A complementary line of work asks whether reliability can be inferred directly from model internals rather than from external judges or repeated sampling. Prior work shows that hidden activations contain useful cues about correctness and hallucination~\citep{burns2022discovering,azaria2023internalstate,zhang2025reasoningmodels}, and recent methods such as \textit{Gnosis}~\citep{ghasemabadi2025can} show that lightweight predictors over internal signals can be effective for scalar correctness estimation. In contrast, we study latent signals as a deployment-time control interface. This setting is broader than scalar reliability estimation and requires decisions over transfer, clarification, tool use, abstention, and direct answering across diverse tasks and environments.

Taken together, these lines of work do not address the central setting of this paper: equipping a frozen LLM or VLM with a practical and lightweight control interface for deployment-time decisions. Our focus is not routing alone, tool use alone, decoding acceleration, or scalar self-verification, but a unified latent control layer for deciding what to do next during agentic inference.

\section{The Multi-Head Latent Control Mechanism}
\vspace{-1em}

In this section, we formalize deployment-time control for frozen foundation models as a latent decision problem over hidden-state trajectories. Building on the overall workflow illustrated in Figure~\ref{fig:main}, we introduce the problem setup, define the latent representations used for control, and describe how these signals are decoded to guide model selection and intervention decisions at inference time.

\subsection{Problem Setup}
\vspace{-0.5em}
Let $x$ denote the LLM input, and let $m_1$ be a frozen primary model that generates an output $\hat{y} = (\hat{y}_1,\ldots,\hat{y}_N)$. When available, we also assume access to a stronger fallback model $m_2$, which can take over when $m_1$ is unlikely to solve the instance reliably. Let $d$ denote the hidden size of $m_1$, and for any layer $\ell$, let $H^{(\ell)} = [h^{(\ell)}_1;\ldots;h^{(\ell)}_N] \in \mathbb{R}^{N \times d}$ denote the sequence of hidden states aligned with the generated output tokens, where $h^{(\ell)}_t$ corresponds to token $\hat{y}_t$.

The two control heads may read different hidden-state traces from the same frozen backbone, since model adequacy and resolution decisions need not be most separable at the same depth. In our default setting, the Capability Head reads the final-layer trace $H^{\mathrm{cap}} = H^{(L)}$, while the Resolution Head reads a selected middle-layer trace $H^{\mathrm{res}} = H^{(\ell_{\mathrm{res}})}$. We compress these variable-length traces into fixed-budget representations $\tilde{H}^{\mathrm{cap}} = \Pi_{\mathrm{cap}}(H^{\mathrm{cap}})$ and $\tilde{H}^{\mathrm{res}} = \Pi_{\mathrm{res}}(H^{\mathrm{res}})$. The layer choices are selected empirically; the corresponding layer-choice ablations for the Capability and Resolution Heads are reported in Appendix~\ref{app:global_layer_ablation} \& ~\ref{app:local_layer_ablation}, respectively.

Multi-head latent control then produces two outputs: a scalar capability score $p_{\mathrm{cap}} \in [0,1]$, which indicates whether control should remain with $m_1$ or be transferred to $m_2$, and a resolution score vector $\mathbf{s}_{\mathrm{res}} = [s_{\mathrm{info}}, s_{\mathrm{tool}}, s_{\mathrm{cant}}] \in [0,1]^3$, which governs the dominant within-model intervention when the instance remains with $m_1$. The backbone is frozen; only the lightweight control heads are trained.

\subsection{Control Prediction from Hidden States}
\vspace{-0.5em}
The two heads decode different control signals from their respective projected traces:
\begin{equation}
z_{\mathrm{cap}} = e_{\phi}^{\mathrm{cap}}(\tilde{H}^{\mathrm{cap}}), \qquad
z_{\mathrm{res}} = e_{\phi}^{\mathrm{res}}(\tilde{H}^{\mathrm{res}}).
\end{equation}
They then predict
\begin{equation}
p_{\mathrm{cap}} = \sigma\!\left(h_{\mathrm{cap}}(z_{\mathrm{cap}})\right), \qquad
\mathbf{s}_{\mathrm{res}} = \sigma\!\left(h_{\mathrm{res}}(z_{\mathrm{res}})\right).
\end{equation}

The \textbf{Capability Head} predicts whether the current model is adequate for the instance or should transfer control to a stronger model, while the \textbf{Resolution Head} determines the appropriate resolution action when computation remains with the current model. This decomposition enables the same latent generation process to support two complementary decisions: instance-level model adequacy and action-level intervention decision.
Both heads operate on hidden-state traces from the model’s completion process, using the same encoder architecture as in~\citep{ghasemabadi2025can}. We restrict inputs to hidden states aligned with generated tokens, excluding the prompt and other conditioning signals, ensuring a uniform interface across text-only and vision-language settings. The encoder is kept fixed to isolate the contribution of our method to the control formulation itself—namely, the capability–resolution decomposition, head-specific supervision, and the resulting inference-time control policy.
\subsection{Inference-Time Policy}
\vspace{-0.5em}
As shown in Figure \ref{fig:main}, the two heads produce control scores at inference time that determine the execution path. We first consult the Capability Head. Larger values of $p_{\mathrm{cap}}$ indicate that the primary model is more likely to be adequate. We therefore hand off to the stornger model $m_2$ only when $p_{\mathrm{cap}} < \tau_{\mathrm{cap}}$; otherwise, the instance remains with $m_1$. When computation stays with $m_1$, the Resolution Head predicts scores over $\mathcal{A} = \{\mathrm{info}, \mathrm{tool}, \mathrm{cant}\}$.
If $\max_{a\in\mathcal{A}_{\mathrm{res}}} s_{\mathrm{res},a}
> \tau_{\mathrm{res}}$, the system executes
$a^\ast=\arg\max_{a\in\mathcal{A}_{\mathrm{res}}} s_{\mathrm{res},a}$;
otherwise, it proceeds directly. The system then requests more information, invokes tools, abstains under the current setup, or proceeds directly when no intervention signal is sufficiently strong. The direct-answer case is represented implicitly by the all-zero resolution state. In this way, the control heads expose an actionable inference-time interface without modifying the backbone or introducing an additional orchestration model. Unless otherwise noted, we report results with a fixed operating point, using a Capability threshold of $0.8$ and triggering a resolution intervention only when the top Resolution score exceeds $0.5$. In Appendix, Figure~\ref{fig:tradeoff_qwen35_qwen3vl} shows that higher routing thresholds hand off more aggressively to the stronger model, increasing performance at higher cost.

\subsection{Control Heads Training}
\vspace{-0.5em}
We freeze the backbone and train only the two lightweight control heads. Although both heads operate on the same latent generation process, they are supervised differently to address distinct deployment-time decisions: the \textbf{Capability Head} predicts instance-level adequacy for transfer, while the \textbf{Resolution Head} predicts structured within-model intervention needs. 
For training, we generate model outputs using the same frozen backbone as at inference time, collect the corresponding hidden-state traces, and derive supervision targets for each head from these outputs. The heads are thus trained to infer the appropriate control decisions directly from latent representations rather than from the model’s surface responses. Crucially, this allows the heads to recover correct control signals even when the model’s final answer or chosen action is incorrect, as such information may still be encoded in the hidden-state trajectory.

\subsubsection{Capability Head}
\vspace{-0.5em}
For the \textbf{Capability Head}, we construct training data by generating responses with the frozen backbone, collecting the corresponding final-layer hidden-state traces, and assigning each instance a scalar adequacy label. The head is trained to predict whether the current model is sufficient for the instance or should transfer control to a stronger model. By default, it operates on the final-layer trace; alternative layer choices are evaluated in Appendix~\ref{app:global_layer_ablation}.

We supervise the head with a scalar adequacy score $y_i^{\mathrm{cap}} \in [0,1]$, generated by an LLM-based judge that compares the model’s output to the reference answer. Because responses may be partially correct or incomplete, scalar supervision provides a more informative signal than binary correctness labels, while aligning directly with the decision of whether the current model is sufficient. Our goal is not to learn a verifier specialized to a narrow domain, but to learn a broadly transferable adequacy signal that remains informative across domains, task types, and modalities. To induce such a signal, we train this head on a deliberately heterogeneous mixture spanning multimodal QA, visual understanding, grounding, long-form reasoning, open-domain QA, API-centric tasks, and more explicitly agentic settings; full data and scoring details are deferred to Appendix~\ref{app:global_data}. 

We also study prefix-trained variants of the Capability Head to investigate how early the adequacy signal becomes reliable. The main prefix-time comparison is reported in Table~\ref{tab:early_global_head}, while a broader prefix-length ablation is deferred to Appendix~\ref{app:prefix_ablation}.

Let $\mathcal{D}_{\mathrm{cap}} = \{(\tilde{H}^{\mathrm{cap}}_i, y_i^{\mathrm{cap}})\}_{i=1}^{M_{\mathrm{cap}}}$ denote the training set. Given the prediction $p_{\mathrm{cap}}^{(i)}$, we optimize
\begin{equation}
\mathcal{L}_{\mathrm{cap}}
=
\frac{1}{M_{\mathrm{cap}}}
\sum_{i=1}^{M_{\mathrm{cap}}}
w_i \, \ell_{\mathrm{reg}} \!\left(p_{\mathrm{cap}}^{(i)}, y_i^{\mathrm{cap}}\right),
\end{equation}
where $\ell_{\mathrm{reg}}$ is a regression loss and $w_i$ reweights examples to compensate for imbalance in the training data, which contains unequal numbers of more-adequate and less-adequate responses. In our default setting, $\ell_{\mathrm{reg}}$ is weighted mean squared error.

\subsubsection{Resolution Head}
\vspace{-0.5em}
The \textbf{Resolution Head} predicts the appropriate intervention when computation remains with the current model. We train it on \textsc{When2Call}, where each instance requires one of four behaviors: answer directly, invoke a tool, request additional information, or abstain under the current setup. We define the explicit action space as $\mathcal{A} = \{\mathrm{info}, \mathrm{tool}, \mathrm{cant}\}$, with direct answer represented implicitly by the all-zero target vector. This design reflects the role of the Resolution Head: it activates only when intervention is required, while direct answering corresponds to the absence of such need.

For each training example, we generate the backbone’s output, collect the corresponding hidden-state trace, and pair it with the correct resolution label. The model’s explicit action is not always correct—for example, it may invoke a tool when clarification is required. The Resolution Head is therefore trained to recover the appropriate decision from the latent trajectory rather than from the model’s surface behavior. Details of label construction are provided in Appendix~\ref{app:when2call_labels}.

Let $\mathcal{D}_{\mathrm{res}} = \{(\tilde{H}^{\mathrm{res}}_i, \mathbf{y}_i^{\mathrm{res}})\}_{i=1}^{M_{\mathrm{res}}}$, where $\mathbf{y}_i^{\mathrm{res}} \in \{0,1\}^3$. Given the prediction $\mathbf{s}_{\mathrm{res}}^{(i)}$, we optimize
\begin{equation}
\mathcal{L}_{\mathrm{res}}
=
\frac{1}{M_{\mathrm{res}}}
\sum_{i=1}^{M_{\mathrm{res}}}
\sum_{a \in \mathcal{A}}
\ell_{\mathrm{BCE}}\!\left(s_a^{(i)}, y_{i,a}^{\mathrm{res}}\right).
\end{equation}

\begin{table*}[t]
\centering
\small
\setlength{\tabcolsep}{2pt}
\renewcommand{\arraystretch}{1.10}
\begin{tabular}{lV cV cccV cccV c}
\toprule
\multirow{2}{*}{\textbf{System}} & \multirow{2}{*}{\textbf{Score}} 
& \multicolumn{3}{cV}{\textbf{$m_1$ Stats}} 
& \multicolumn{3}{cV}{\textbf{$m_2$ Stats}} 
& \multirow{2}{*}{\textbf{API Cost}} \\
& & \textbf{In Tok.} & \textbf{Out Tok.} & \textbf{\#Calls}
  & \textbf{In Tok.} & \textbf{Out Tok.} & \textbf{\#Calls}
  & \\
\midrule

$m_1$: Qwen3-VL-4B & 0.47
& 14.15M & 0.47M & 3250
& 0 & 0 & 0
& 0 \\

$m_2$: Qwen3-VL-32B & 0.58
& 0 & 0 & 0
& 14.74M & 0.53M & 3595
& \$2.70 \\

\rowcolor{blue!8}
Routed: Qwen3-VL-4B $\rightarrow$ 32B & 0.60
& 15.14M & 0.56M & 3667
& 1.39M & 0.05M & 300
& \$0.25 {\color{green!50!black}(\(\downarrow\) 90.7\%)} \\

\midrule
$m_1$: Qwen3.5-9B & 0.51
& 9.86M & 0.36M & 2282
& 0 & 0 & 0
& 0 \\

$m_2$: Qwen3.5-27B & 0.59
& 0 & 0 & 0
& 12.75M & 0.57M & 2854
& \$1.49 \\

\rowcolor{blue!8}
Routed: Qwen3.5-9B $\rightarrow$ 27B & 0.56
& 10.2M & 0.37M & 2170
& 1.9M & 0.07M & 470
& \$0.21 {\color{green!50!black}(\(\downarrow\) 85.8\%)} \\

\bottomrule
\end{tabular}
\caption{AndroidWorld success rate and backend usage for single-model and routed runs. The smaller model is assumed to run locally for free, so reported API cost reflects only large-model usage. Dynamic Routed execution improves success while sharply reducing paid large-model calls.}
\vspace{-1.5em}
\label{tab:androidworld_backend}
\end{table*}
\vspace{-1em}
\section{Experiments and Results}
\vspace{-1em}

We evaluate Multi-Head Latent Control across two complementary dimensions: \emph{capability prediction} and \emph{resolution decision-making}, each designed to test a distinct aspect of deployment-time control.
For \textbf{capability-guided routing} (\S\ref{sec:result_capability}), we study routed collaboration between a smaller primary model and a stronger fallback model across multiple backbone families and benchmark suites, focusing on score--cost tradeoffs. We further include AndroidWorld as a long-horizon agentic case study to assess real-world cost and usage behavior under sequential decision-making.
For \textbf{resolution decision-making} (\S\ref{sec:result_resolution}), we evaluate the Resolution Head on \textsc{When2Call}, where each instance requires selecting among clarification, tool use, abstention, or direct answering. This setting isolates the quality of intervention decisions by comparing the backbone’s native behavior against the same model augmented with our control layer.
We also study two deployment-oriented variants: (i) a web-augmented TriviaQA setting (\S\ref{sec:result_trivia}) to evaluate tool-escalation decisions under incomplete knowledge, and (ii) a prefix-time setting (\S\ref{sec:result_prefix}) to test whether adequacy signals can be recovered from partial generations. Additional ablations on layer choice, prefix length, and training mixture are provided in \S\ref{sec:result_ablation}.
Across all settings, we assess whether lightweight heads attached to frozen backbones produce control signals with direct system value, including improved score--cost tradeoffs, more accurate intervention decisions, better tool-escalation behavior, and reliable early adequacy prediction.
\newcommand{\famcell}[2]{\shortstack[l]{$m_1$: #1\\$m_2$: #2}}
\newcommand{\savepct}[1]{{\scriptsize\textcolor{green!60!black}{\(\downarrow\)#1}}}

\begin{table*}[t]
\centering
\small
\setlength{\tabcolsep}{3.0pt}
\renewcommand{\arraystretch}{1.10}
\resizebox{\textwidth}{!}{%
\begin{tabular}{lV lV cV cV cV cV cV cV c}
\toprule
\multirow{2}{*}{\textbf{Family}} & \multirow{2}{*}{\textbf{Setting}}
& \textbf{CharXiv}
& \textbf{MathVerse}
& \textbf{MathVista}
& \textbf{ScreenSpot-Pro}
& \textbf{SimpleVQA}
& \textbf{MMLU-Pro}
& \textbf{Overall} \\
& & \textit{score / cost} & \textit{score / cost} & \textit{score / cost} & \textit{score / cost} & \textit{score / cost} & \textit{score / cost} & \textit{score / cost} \\
\midrule

\multirow{4}{*}{\famcell{Qwen3.5-4B}{Qwen3.5-27B-Thk}}
& $m_1$: & \sblank{0.63} & \sblank{0.86} & \sblank{0.84} & \sblank{0.36} & \sblank{0.37} & \sblank{0.68} & \sblank{0.63} \\
& $m_2$: & \scell{0.73}{\$3.50} & \scell{0.89}{\$4.66} & \scell{0.87}{\$3.06} & \scell{0.65}{\$7.59} & \scell{0.42}{\$2.68} & \scell{0.78}{\$5.88} & \scell{0.72}{\$27.37} \\
& \multirow{2}{*}{Routed}
& \cellcolor{blue!8}\scell{0.70}{\$2.09}
& \cellcolor{blue!8}\scell{0.90}{\$1.31}
& \cellcolor{blue!8}\scell{0.86}{\$1.30}
& \cellcolor{blue!8}\scell{0.64}{\$5.79}
& \cellcolor{blue!8}\scell{0.41}{\$1.48}
& \cellcolor{blue!8}\scell{0.78}{\$3.21}
& \cellcolor{blue!8}\scell{0.72}{\$15.17} \\
&
& \cellcolor{blue!8}\savepct{40.3\%}
& \cellcolor{blue!8}\savepct{71.9\%}
& \cellcolor{blue!8}\savepct{57.5\%}
& \cellcolor{blue!8}\savepct{23.7\%}
& \cellcolor{blue!8}\savepct{44.8\%}
& \cellcolor{blue!8}\savepct{45.4\%}
& \cellcolor{blue!8}\savepct{44.6\%} \\
\arrayrulecolor{gray!60}\specialrule{0.15pt}{0pt}{1pt}\arrayrulecolor{black}

\multirow{4}{*}{\famcell{Qwen3.5-9B}{Qwen3.5-27B-Thk}}
& $m_1$: & \sblank{0.68} & \sblank{0.88} & \sblank{0.85} & \sblank{0.47} & \sblank{0.41} & \sblank{0.71} & \sblank{0.67} \\
& $m_2$: & \scell{0.73}{\$3.50} & \scell{0.89}{\$4.66} & \scell{0.87}{\$3.06} & \scell{0.65}{\$7.59} & \scell{0.42}{\$2.68} & \scell{0.80}{\$5.85} & \scell{0.73}{\$27.35} \\
& \multirow{2}{*}{Routed}
& \cellcolor{blue!8}\scell{0.73}{\$2.48}
& \cellcolor{blue!8}\scell{0.91}{\$1.24}
& \cellcolor{blue!8}\scell{0.87}{\$1.00}
& \cellcolor{blue!8}\scell{0.64}{\$4.57}
& \cellcolor{blue!8}\scell{0.42}{\$1.25}
& \cellcolor{blue!8}\scell{0.78}{\$2.30}
& \cellcolor{blue!8}\scell{0.72}{\$12.85} \\
&
& \cellcolor{blue!8}\savepct{29.1\%}
& \cellcolor{blue!8}\savepct{73.4\%}
& \cellcolor{blue!8}\savepct{67.3\%}
& \cellcolor{blue!8}\savepct{39.8\%}
& \cellcolor{blue!8}\savepct{53.4\%}
& \cellcolor{blue!8}\savepct{60.7\%}
& \cellcolor{blue!8}\savepct{53.0\%} \\
\arrayrulecolor{gray!60}\specialrule{0.15pt}{0pt}{1pt}\arrayrulecolor{black}

\multirow{4}{*}{\famcell{Qwen3-VL-2B-Thk}{Qwen3-VL-32B-Thk}}
& $m_1$: & \sblank{0.38} & \sblank{0.74} & \sblank{0.72} & \sblank{0.27} & \sblank{0.33} & \sblank{0.54} & \sblank{0.50} \\
& $m_2$: & \scell{0.64}{\$1.33} & \scell{0.86}{\$2.75} & \scell{0.84}{\$1.42} & \scell{0.52}{\$6.48} & \scell{0.41}{\$0.68} & \scell{0.72}{\$2.53} & \scell{0.67}{\$15.19} \\
& \multirow{2}{*}{Routed}
& \cellcolor{blue!8}\scell{0.62}{\$1.19}
& \cellcolor{blue!8}\scell{0.85}{\$1.45}
& \cellcolor{blue!8}\scell{0.84}{\$0.96}
& \cellcolor{blue!8}\scell{0.49}{\$4.84}
& \cellcolor{blue!8}\scell{0.39}{\$0.40}
& \cellcolor{blue!8}\scell{0.72}{\$2.22}
& \cellcolor{blue!8}\scell{0.65}{\$11.06} \\
&
& \cellcolor{blue!8}\savepct{10.5\%}
& \cellcolor{blue!8}\savepct{47.3\%}
& \cellcolor{blue!8}\savepct{32.4\%}
& \cellcolor{blue!8}\savepct{25.3\%}
& \cellcolor{blue!8}\savepct{41.2\%}
& \cellcolor{blue!8}\savepct{12.3\%}
& \cellcolor{blue!8}\savepct{27.2\%} \\
\arrayrulecolor{gray!60}\specialrule{0.15pt}{0pt}{1pt}\arrayrulecolor{black}

\multirow{4}{*}{\famcell{Qwen3-VL-4B}{Qwen3-VL-32B-Thk}}
& $m_1$: & \sblank{0.41} & \sblank{0.79} & \sblank{0.78} & \sblank{0.54} & \sblank{0.35} & \sblank{0.63} & \sblank{0.58} \\
& $m_2$: & \scell{0.64}{\$1.33} & \scell{0.86}{\$2.75} & \scell{0.84}{\$1.42} & \scell{0.52}{\$6.48} & \scell{0.41}{\$0.68} & \scell{0.73}{\$2.52} & \scell{0.67}{\$15.17} \\
& \multirow{2}{*}{Routed}
& \cellcolor{blue!8}\scell{0.63}{\$1.16}
& \cellcolor{blue!8}\scell{0.86}{\$1.49}
& \cellcolor{blue!8}\scell{0.83}{\$0.70}
& \cellcolor{blue!8}\scell{0.57}{\$2.62}
& \cellcolor{blue!8}\scell{0.39}{\$0.43}
& \cellcolor{blue!8}\scell{0.71}{\$1.58}
& \cellcolor{blue!8}\scell{0.67}{\$7.98} \\
&
& \cellcolor{blue!8}\savepct{12.8\%}
& \cellcolor{blue!8}\savepct{45.8\%}
& \cellcolor{blue!8}\savepct{50.7\%}
& \cellcolor{blue!8}\savepct{59.6\%}
& \cellcolor{blue!8}\savepct{36.8\%}
& \cellcolor{blue!8}\savepct{37.3\%}
& \cellcolor{blue!8}\savepct{47.4\%} \\
\arrayrulecolor{gray!60}\specialrule{0.15pt}{0pt}{1pt}\arrayrulecolor{black}

\multirow{4}{*}{\famcell{Qwen3-VL-4B-Thk}{Qwen3-VL-32B-Thk}}
& $m_1$: & \sblank{0.53} & \sblank{0.82} & \sblank{0.78} & \sblank{0.37} & \sblank{0.37} & \sblank{0.65} & \sblank{0.59} \\
& $m_2$: & \scell{0.64}{\$1.33} & \scell{0.86}{\$2.75} & \scell{0.84}{\$1.42} & \scell{0.52}{\$6.48} & \scell{0.41}{\$0.68} & \scell{0.73}{\$2.47} & \scell{0.67}{\$15.13} \\
& \multirow{2}{*}{Routed}
& \cellcolor{blue!8}\scell{0.63}{\$0.88}
& \cellcolor{blue!8}\scell{0.86}{\$0.98}
& \cellcolor{blue!8}\scell{0.84}{\$0.92}
& \cellcolor{blue!8}\scell{0.49}{\$3.88}
& \cellcolor{blue!8}\scell{0.41}{\$0.31}
& \cellcolor{blue!8}\scell{0.72}{\$1.43}
& \cellcolor{blue!8}\scell{0.66}{\$8.40} \\
&
& \cellcolor{blue!8}\savepct{33.8\%}
& \cellcolor{blue!8}\savepct{64.4\%}
& \cellcolor{blue!8}\savepct{35.2\%}
& \cellcolor{blue!8}\savepct{40.1\%}
& \cellcolor{blue!8}\savepct{54.4\%}
& \cellcolor{blue!8}\savepct{42.1\%}
& \cellcolor{blue!8}\savepct{44.5\%} \\
\arrayrulecolor{gray!60}\specialrule{0.15pt}{0pt}{1pt}\arrayrulecolor{black}

\multirow{4}{*}{\famcell{Gemma-4B}{Gemma-31B-Thk}}
& $m_1$: & \sblank{0.41} & \sblank{0.65} & \sblank{0.69} & -- & \sblank{0.30} & \sblank{0.61} & \sblank{0.53} \\
& $m_2$: & \scell{0.63}{\$0.62} & \scell{0.87}{\$1.34} & \scell{0.80}{\$0.81} & -- & \scell{0.41}{\$0.39} & \scell{0.78}{\$1.05} & \scell{0.70}{\$4.20} \\
& \multirow{2}{*}{Routed}
& \cellcolor{blue!8}\scell{0.58}{\$0.49}
& \cellcolor{blue!8}\scell{0.87}{\$1.02}
& \cellcolor{blue!8}\scell{0.79}{\$0.62}
& -- 
& \cellcolor{blue!8}\scell{0.41}{\$0.24}
& \cellcolor{blue!8}\scell{0.75}{\$0.68}
& \cellcolor{blue!8}\scell{0.68}{\$3.05} \\
&
& \cellcolor{blue!8}\savepct{21.0\%}
& \cellcolor{blue!8}\savepct{23.9\%}
& \cellcolor{blue!8}\savepct{23.5\%}
& --
& \cellcolor{blue!8}\savepct{38.5\%}
& \cellcolor{blue!8}\savepct{35.2\%}
& \cellcolor{blue!8}\savepct{27.4\%} \\
\arrayrulecolor{gray!60}\specialrule{0.15pt}{0pt}{1pt}\arrayrulecolor{black}

\multirow{4}{*}{\famcell{Gemma-4B-Thk}{Gemma-31B-Thk}}
& $m_1$: & \sblank{0.42} & \sblank{0.65} & \sblank{0.67} & -- & \sblank{0.31} & \sblank{0.62} & \sblank{0.53} \\
& $m_2$: & \scell{0.63}{\$0.62} & \scell{0.87}{\$1.34} & \scell{0.80}{\$0.81} & -- & \scell{0.41}{\$0.39} & \scell{0.78}{\$1.05} & \scell{0.70}{\$4.21} \\
& \multirow{2}{*}{Routed}
& \cellcolor{blue!8}\scell{0.59}{\$0.43}
& \cellcolor{blue!8}\scell{0.84}{\$1.02}
& \cellcolor{blue!8}\scell{0.76}{\$0.44}
& --
& \cellcolor{blue!8}\scell{0.41}{\$0.29}
& \cellcolor{blue!8}\scell{0.76}{\$0.71}
& \cellcolor{blue!8}\scell{0.67}{\$2.89} \\
&
& \cellcolor{blue!8}\savepct{30.6\%}
& \cellcolor{blue!8}\savepct{23.9\%}
& \cellcolor{blue!8}\savepct{45.7\%}
& --
& \cellcolor{blue!8}\savepct{25.6\%}
& \cellcolor{blue!8}\savepct{32.4\%}
& \cellcolor{blue!8}\savepct{31.4\%} \\
\bottomrule
\end{tabular}}
\caption{Score--cost tradeoff across model families and benchmarks. $m_1$ is the local model, $m_2$ the always-large baseline, and Routed switches between them using the Capability Head. Each cell reports score and estimated paid API cost. Green indicates reduction in large-model cost relative to the $m_2$ baseline; $m_1$ is assumed to run locally at no cost.}
\vspace{-2em}
\label{tab:main_score_cost}
\end{table*}
\vspace{-0.8em}
\paragraph{Backbones.}
We evaluate three backbone families: \textbf{Qwen3-VL}~\citep{qwen3vl2025}, \textbf{Qwen3.5}~\citep{qwen352026}, and \textbf{Gemma}~\citep{gemma42026}. Across these families, our routed systems span model sizes from 2B to 32B and include both thinking and non-thinking variants, allowing us to test whether the proposed control layer transfers across different model lines, scales, and inference modes rather than a single narrow setup.
\vspace{-0.8em}
\paragraph{Training/eval pipeline and infrastructure.}
We train lightweight control heads on top of Qwen3-VL-2B/4B/32B, Qwen3-VL-2B-Thk/4B-Thk, Qwen3.5-4B/9B, and Gemma-4B/4B-Thk backbones, \textit{Thk} denotes thinking mode. Training labels are constructed by comparing each model generation with the ground-truth answer using Qwen3vl 30B-A3B as judge model. We optimize the heads with Adam at a learning rate of $1\times10^{-4}$. As backbone remains frozen, the full pipeline is lightweight: in a representative 9B-scale Qwen setting, data generation and training for both heads completes in under one day on a single high-end GPU, and head training itself fits within 16\,GB of GPU memory. 
\vspace{-0.8em}
\paragraph{Benchmarks.}
We evaluate across a deliberately diverse set of tasks and deployment settings rather than a single benchmark type. For capability collaboration, we report results on \textbf{SimpleVQA}~\citep{cheng2025simplevqa}, \textbf{ScreenSpot-Pro}~\citep{li2025screenspotpro}, \textbf{CharXiv-Reasoning}~\citep{wang2024charxiv}, \textbf{MathVerse}~\citep{zhang2024mathverse}, \textbf{MathVista}~\citep{lu2024mathvista}, and \textbf{MMLU-Pro}~\citep{wang2024mmlupro}, covering multimodal perception, GUI grounding, chart and document reasoning, mathematical reasoning, and broad knowledge tasks. We additionally include \textbf{AndroidWorld}~\citep{rawles2024androidworld} as a long-horizon, multi-step agentic case study with backend usage statistics, implemented using the Mobile-Agent-v3.5 framework~\citep{xu2026mobile}. 
For resolution control, we evaluate structured intervention prediction on \textsc{When2Call}~\citep{ross2025when2call}. Finally, we study \textbf{TriviaQA}~\citep{joshi2017triviaqa} as a concrete external-action setting for web-search decision quality. Taken together, these benchmarks test generalization across backbone families, task formats, modalities, and deployment behaviors.
\vspace{-0.8em}

\paragraph{Metrics.}
For multi-agent experiments, we report each benchmark’s native task score alongside estimated API cost. In Table~\ref{tab:main_score_cost}, each entry shows \textit{score / estimated paid cost}, with the Overall column reporting average score and total cost across benchmarks. In Table~\ref{tab:androidworld_backend}, we further report backend usage statistics—including input tokens, output tokens, and number of calls for both primary and fallback models—to characterize changes in large-model usage under routed execution.

In settings where we want to compare the quality of the Capability signal directly, we report ROC-AUC, AUPR-C, AUPR-I, and ECE. ROC-AUC measures how well the adequacy signal ranks more-adequate above less-adequate cases, AUPR-C and AUPR-I measure precision--recall quality under the two complementary positive classes, and ECE measures calibration. For the Resolution Head on \textsc{When2Call}, Table~\ref{tab:when2call_main} reports F1 score and accuracy. For the TriviaQA web-search analysis, Table~\ref{tab:web_overuse} reports model's score together with the number of web calls, tool-call precision, and missed-needed web calls. Estimated API cost is reported as a deployment-oriented proxy based on token usage; full details are deferred to Appendix~\ref{app:cost_estimation}.
\vspace{-0.1cm}
\vspace{-0.5em}
\subsection{Efficient Multi-Model Collaboration}\label{sec:result_capability}
\vspace{-0.5em}
We evaluate the \textbf{Capability Head} in the primary deployment setting: collaboration between a weaker local model and a stronger fallback model. Table~\ref{tab:main_score_cost} reports score--cost tradeoffs across benchmarks and model families, while Table~\ref{tab:androidworld_backend} presents AndroidWorld as a long-horizon agentic case study. The key question is whether a lightweight adequacy signal can recover the benefits of a stronger model without incurring its cost on every instance. 
Across settings, routed systems consistently outperform the smaller-model baselines while remaining substantially cheaper than always using the stronger model. For example, routed Qwen3.5 systems reduce paid cost by approximately 45--53\% while retaining most of the large-model performance, and routed Qwen3-VL systems achieve similar savings with comparable performance preservation.

AndroidWorld highlights this effect clearly. It is a long-horizon benchmark where the model is invoked repeatedly over multi-step trajectories involving planning, acting, tool use, and reflection. In our routed setup, the Capability Head decides at each step whether to remain local or escalate control, rather than relying on a large model throughout or fixed role assignment. 
This yields substantial savings while preserving or improving performance: \textit{Qwen3-VL-4B $\rightarrow$ 32B} improves score from 0.47 to 0.60 with a 90.7\% reduction in paid API cost, and \textit{Qwen3.5-9B $\rightarrow$ 27B} improves score from 0.51 to 0.56 with an 85.8\% cost reduction.

Appendix~\ref{app:self_switch_prompt_baseline} and Appendix~\ref{app:confidence_vs_head_screenspot} evaluate two natural alternatives to latent-control routing. The first considers a self-abstention strategy, where the smaller model is prompted to defer when uncertain; however, this leads to severe under-escalation and fails to recover the performance gains of latent-control routing on ScreenSpot-Pro and MMLU-Pro. The second examines token-level confidence as a proxy for adequacy: on Qwen3.5-9B over ScreenSpot-Pro, confidence scores poorly distinguish correct from incorrect predictions, whereas the Capability Head provides a significantly more discriminative signal.

\vspace{-0.5em}
\subsection{Resolution Control on \textsc{When2Call}}\label{sec:result_resolution}
\vspace{-0.5em}
We next isolate the {Resolution Head} to evaluate whether hidden-state representations can reliably support within-model intervention decisions.

Table~\ref{tab:when2call_main} evaluates the Resolution Head on \textsc{When2Call} across several frozen backbones. For each model, we compare the backbone's native action choice, obtained from its surface response, against the same backbone augmented with the Resolution Head. We also include the task-specialized \textsc{When2Call} baselines reported by the benchmark as reference points. Unlike our lightweight head-based adaptation, these baselines are full LLMs fine-tuned directly on \textsc{When2Call}.

The main result is that latent control consistently improves resolution decision quality over the backbone's explicit behavior. The gains are often substantial: for example, on Qwen-VL-2B, F1 improves from 37.3 to 49.0 and accuracy from 52.7 to 65.1, while on Qwen3.5-4B, F1 improves from 43.5 to 54.5 and accuracy from 57.9 to 69.5. Even when the model's surface action is wrong, its hidden-state contains information for the head to recover the correct resolution decision. These results show that lightweight control heads can improve structured intervention decisions without modifying the backbone.

\begin{table*}[t]
\centering
\scriptsize
\setlength{\tabcolsep}{0.8pt}
\renewcommand{\arraystretch}{1.06}
\resizebox{\textwidth}{!}{%
\begin{tabular}{lV ccV ccV ccV ccV ccV cc}
\toprule
\multirow{2}{*}{\textbf{Method}} 
& \multicolumn{2}{cV}{\textbf{Qwen-VL-2B}} 
& \multicolumn{2}{cV}{\textbf{Qwen-VL-4B}} 
& \multicolumn{2}{cV}{\textbf{Qwen-VL-2B-Thk}} 
& \multicolumn{2}{cV}{\textbf{Qwen3.5-4B}}
& \multicolumn{2}{cV}{\textbf{Gemma-2B}} 
& \multicolumn{2}{c}{\textbf{Gemma-2B-Thk}} \\
& \textbf{F1} & \textbf{Acc}
& \textbf{F1} & \textbf{Acc}
& \textbf{F1} & \textbf{Acc}
& \textbf{F1} & \textbf{Acc}
& \textbf{F1} & \textbf{Acc}
& \textbf{F1} & \textbf{Acc} \\
\midrule
\rowcolor{gray!6}
MNM 4B When2Call-SFT \cite{ross2025when2call}
& \multicolumn{12}{c}{F1 = 48.1 \quad Acc = 67.8} \\

\rowcolor{gray!6}
MNM 8B When2Call-SFT \cite{ross2025when2call}
& \multicolumn{12}{c}{F1 = 49.4 \quad Acc = 68.2} \\
\midrule

Backbone Choice
& 37.3 & 52.7
& 48.7 & 63.9
& 55.1 & 73.6
& 43.5 & 57.9
& 40.1 & 56.5
& 54.2 & 70.1 \\

\rowcolor{blue!8}
Backbone + Resolution Head
& 49.0 & 65.1
& 52.7 & 70.1
& 54.5 & 69.5
& 50.0 & 63.6
& 55.2 & 71.2 \\
\bottomrule
\end{tabular}%
}
\caption{Resolution decision prediction on \textsc{When2Call}. We compare each frozen backbone's native action choice against the same backbone augmented with the Resolution Head; First two rows show full-LLM \textsc{When2Call} baselines fine-tuned directly on the benchmark.}
\vspace{-1em}
\label{tab:when2call_main}
\end{table*}




\vspace{-0.5em}
\subsection{Web-Tool Decision Quality on TriviaQA}\label{sec:result_trivia}
\vspace{-0.5em}
We next study a concrete deployment variant of Capability Head: using the Capability Head to decide whether the current model should answer locally or escalate to web search. TriviaQA is a useful testbed for this setting because retrieval is often helpful, but not every instance requires an external search. Table~\ref{tab:web_overuse} compares the backbone's native web-search behavior against the same backbone augmented with the Capability Head. Here, the control signal is used in the same spirit as routed collaboration, except that inadequate cases are escalated to a tool call rather than to a stronger fallback model. The question is therefore not whether the head simply suppresses tool use, but whether it improves the quality of the decision to use external retrieval.

Across backbones, the Capability Head consistently improves task score and reduces missed-needed web calls, indicating better recovery of when web search is actually required. In several cases, this comes with more web calls rather than fewer, so the main effect is not blanket suppression of tool use. Instead, the adequacy signal improves the quality of the escalation decision. This effect is visible even for very strong large models: on \textit{Qwen3-VL-32B}, adding the control head improves tool-call precision from 72.7\% to 75.5\%, raises task score from 0.672 to 0.778, and reduces missed-needed web calls from 328 to 222.

\begin{table*}[t]
\centering
\small
\setlength{\tabcolsep}{1.2pt}
\renewcommand{\arraystretch}{1.03}
\begin{tabular}{lV cV cV cV cV cV cV cV cV c}
\toprule
& & \multicolumn{4}{cV}{\textbf{Backbone Choice}} 
& \multicolumn{4}{c}{\textbf{Backbone + Capability Head}} \\
\cmidrule(lr){3-6} \cmidrule(lr){7-10}
\textbf{Backbone} &
{\scriptsize \textbf{No-Tool} $\uparrow$} &
{\scriptsize \textbf{Score} $\uparrow$} &
{\scriptsize \textbf{Calls}} &
{\scriptsize \textbf{Precision (\%)} $\uparrow$} &
{\scriptsize \textbf{Missed} $\downarrow$} &
{\scriptsize \textbf{Score} $\uparrow$} &
{\scriptsize \textbf{Calls}} &
{\scriptsize \textbf{Precision (\%)} $\uparrow$} &
{\scriptsize \textbf{Missed} $\downarrow$} \\
\midrule
Qwen3-VL-4B  & 0.231 & 0.292 & 67  & 91.0 & 708
& \cellcolor{blue!8} 0.756 {\tiny\color{green!50!black}(+158.9\%)}
& \cellcolor{blue!8} 518 {\tiny(+451)}
& \cellcolor{blue!8} 87.1 {\tiny\color{red!70!black}(-3.9)}
& \cellcolor{blue!8} 244 {\tiny\color{green!50!black}(-464)} \\

Qwen3-VL-32B & 0.656 & 0.672 & 22 & 72.7 & 328
& \cellcolor{blue!8} 0.778 {\tiny\color{green!50!black}(+15.8\%)}
& \cellcolor{blue!8} 163 {\tiny(+141)}
& \cellcolor{blue!8} 75.5 {\tiny\color{green!50!black}(+2.8)}
& \cellcolor{blue!8} 222 {\tiny\color{green!50!black}(-106)} \\

Qwen3.5-9B   & 0.593 & 0.624 & 39  & 79.5 & 376
& \cellcolor{blue!8} 0.858 {\tiny\color{green!50!black}(+37.5\%)}
& \cellcolor{blue!8} 357 {\tiny(+318)}
& \cellcolor{blue!8} 75.4 {\tiny\color{red!70!black}(-4.1)}
& \cellcolor{blue!8} 142 {\tiny\color{green!50!black}(-234)} \\

Gemma-4B     & 0.427 & 0.862 & 608 & 71.5 & 138
& \cellcolor{blue!8} 0.921 {\tiny\color{green!50!black}(+6.8\%)}
& \cellcolor{blue!8} 620 {\tiny(+12)}
& \cellcolor{blue!8} 77.1 {\tiny\color{green!50!black}(+5.6)}
& \cellcolor{blue!8} 79 {\tiny\color{green!50!black}(-59)} \\

Gemma-4B-Thk & 0.447 & 0.811 & 457 & 79.6 & 189
& \cellcolor{blue!8} 0.902 {\tiny\color{green!50!black}(+11.2\%)}
& \cellcolor{blue!8} 570 {\tiny(+113)}
& \cellcolor{blue!8} 80.7 {\tiny\color{green!50!black}(+1.1)}
& \cellcolor{blue!8} 98 {\tiny\color{green!50!black}(-91)} \\
\bottomrule
\end{tabular}
\caption{TriviaQA web-search decision quality. We compare each backbone's native web-search behavior against a Capability-Head variant that triggers web search only when the model is unlikely to answer correctly from parametric knowledge alone.}
\vspace{-0.5em}
\label{tab:web_overuse}
\end{table*}
\vspace{-0.5em}
\subsection{Prefix-Time Capability Prediction}\label{sec:result_prefix}
\vspace{-0.5em}
The default collaboration setup waits for the weaker model to finish its response before applying the Capability Head. While effective, this can still waste compute when the local trajectory is already heading toward failure. We therefore ask whether the adequacy signal can be recovered earlier from only a short prefix of the generated answer. Table~\ref{tab:early_global_head} evaluates the quality of the Capability signal itself under three regimes: \textit{Full / Full}, \textit{Full / Prefix-200}, and \textit{Prefix-200 / Prefix-200}. As expected, prefix-time prediction is weaker than full-trajectory prediction, since the head observes only a partial hidden-state trace. However, training directly in the prefix regime recovers substantially stronger prefix-time signal quality than naively applying a full-trajectory-trained head to partial generations.

This signal-level difference also translates to end-to-end routed behavior. Appendix Table~\ref{tab:appendix_prefix_routing_tradeoff} shows that training directly on prefixes better recovers the routed performance of full-trajectory control than simply applying a full-trained head at 200 tokens, while still preserving substantial reductions in paid API cost relative to always using the large model. These results show that model adequacy can often be detected early enough to support useful routed inference, and that prefix-time control is best learned in the same regime in which it is applied.

\begin{table*}[t]
\centering
\small
\setlength{\tabcolsep}{2.4pt}
\renewcommand{\arraystretch}{1.08}
\begin{tabular}{ll|cccc|cccc|cccc}
\toprule
\multirow{2}{*}{\textbf{Train}} & \multirow{2}{*}{\textbf{Inference}} 
& \multicolumn{4}{c|}{\textbf{Qwen3-VL-2B-Thk}} 
& \multicolumn{4}{c|}{\textbf{Qwen3-VL-4B-Thk}} 
& \multicolumn{4}{c}{\textbf{Gemma-4B-Thk}} \\
\cmidrule(lr){3-6}\cmidrule(lr){7-10}\cmidrule(lr){11-14}
& & \multicolumn{12}{c}{\textbf{ROC-AUC} $\uparrow$\hspace{0.8em}\textbf{AUPR-C} $\uparrow$\hspace{0.8em}\textbf{AUPR-I} $\uparrow$\hspace{0.8em}\textbf{ECE} $\downarrow$} \\
\midrule
Full & Full
& \textbf{0.85} & \textbf{0.73} & \textbf{0.87} & \textbf{0.20}
& \textbf{0.86} & \textbf{0.91} & \textbf{0.75} & \textbf{0.14}
& \textbf{0.84} & \textbf{0.74} & \textbf{0.88} & \textbf{0.16} \\
Full & Prefix-200
& 0.77 & 0.65 & 0.81 & 0.33
& 0.78 & 0.80 & 0.70 & 0.20
& 0.78 & 0.69 & 0.81 & 0.24 \\
Prefix-200 & Prefix-200
& 0.80 & 0.69 & 0.82 & \textbf{0.20}
& 0.80 & 0.82 & 0.73 & 0.18
& 0.82 & 0.71 & 0.84 & \textbf{0.16} \\
\bottomrule
\end{tabular}
\caption{Prefix-time evaluation of the Capability Head on top of each backbone. Metrics are averaged over the benchmarks in Table~\ref{tab:main_score_cost}. We measure the quality of the adequacy signal itself rather than end-to-end routed performance. Applying a head trained on full completions to prefixes degrades performance, while training directly on prefixes recovers stronger prefix-time quality.}
\vspace{-1em}
\label{tab:early_global_head}
\end{table*}

\vspace{-0.5em}
\subsection{Ablations}\label{sec:result_ablation}
\vspace{-0.5em}
Additional analyses in the appendix support the main design choices of the method. Appendix~\ref{app:global_layer_ablation} ablates the \textbf{hidden-state layer used by the Capability Head} and shows that the \textbf{final layer} provides the \textbf{strongest overall adequacy signal}. Appendix~\ref{app:local_layer_ablation} performs the corresponding study for the \textbf{Resolution Head} and shows that a \textbf{middle-layer trace} is most effective for \textbf{resolution decision prediction}. Appendix~\ref{app:training_mix_ablation} tests the \textbf{breadth of the capability-head training mixture} and shows that \textbf{narrow visual-math-only training transfers substantially worse} than the \textbf{full mixed-data setting} on \textbf{ScreenSpot Pro}, supporting our goal of learning a \textbf{broadly transferable adequacy signal}. Appendix~\ref{app:prefix_ablation} studies how the \textbf{Capability signal changes with available prefix length}, while Table~\ref{tab:appendix_prefix_routing_tradeoff} reports the corresponding \textbf{end-to-end routed score--cost tradeoff} under the same \textbf{prefix-time regimes}. Appendix~\ref{app:self_switch_prompt_baseline} includes a \textbf{prompt-level self-switching baseline}, showing that the model \textbf{rarely escalates to the stronger model when needed} compared with the \textbf{Capability Head}.
\vspace{-0.5em}
\section{Conclusion}
\vspace{-1em}
We introduced multi-head latent control, a lightweight deployment-time control layer that equips frozen foundation models with a practical control interface for deciding what to do next during inference. Rather than modifying the backbone, the method trains small control heads that read hidden-state traces to predict model adequacy and resolution decisions. Across multi-model collaboration, long-horizon agentic execution, structured resolution decision-making, tool-use decisions, and prefix-time prediction, the results show that these signals can improve deployment behavior while preserving the efficiency and reusability of frozen models. More broadly, this suggests that hidden-state traces provide a scalable substrate for lightweight control interfaces that can be rapidly attached to new foundation models as they are deployed.

\bibliographystyle{abbrvnat}
\bibliography{refs}

\newpage
\appendix

\begin{figure*}[t]
    \centering
    \includegraphics[width=\textwidth]{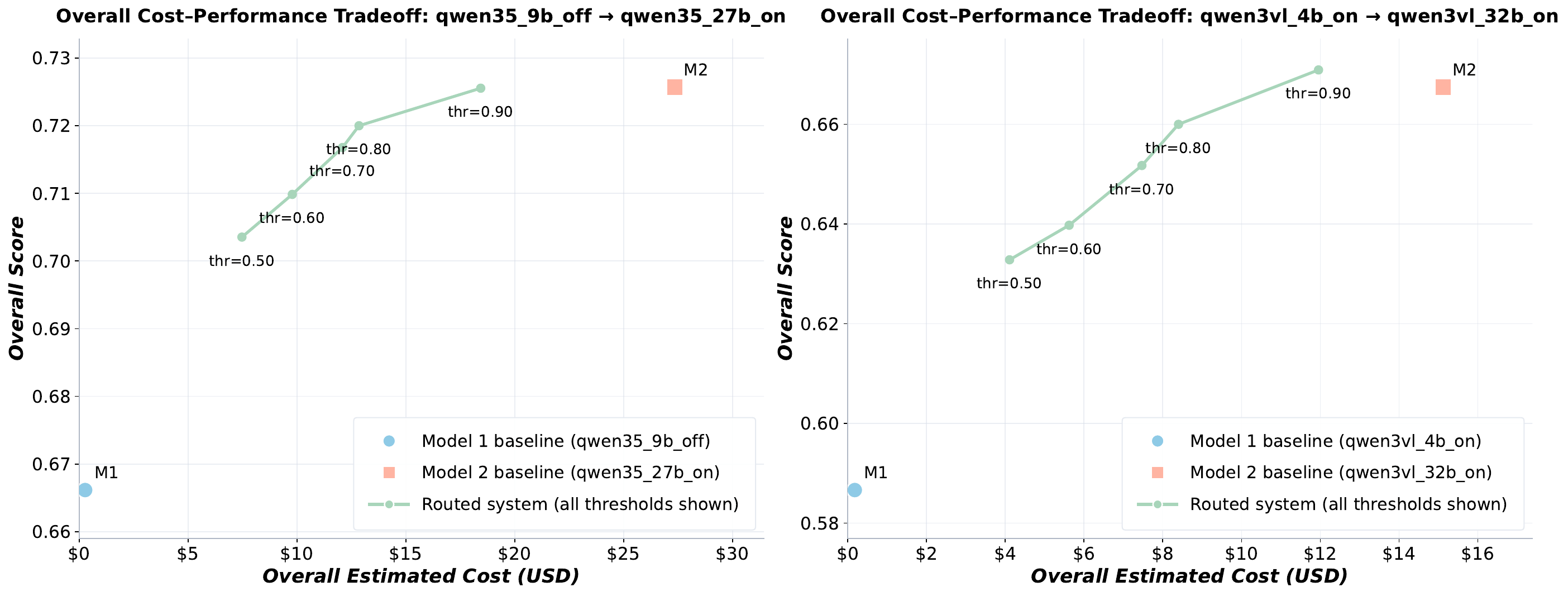}
    \caption{
    Overall cost--performance tradeoff for two routed systems: \textbf{Qwen3.5-9B $\rightarrow$ Qwen3.5-27B-Thk} (left) and \textbf{Qwen3-VL-4B-Thk $\rightarrow$ Qwen3-VL-32B-Thk} (right). 
    Each panel compares the local small model ($m_1$:), the always-large baseline ($m_2$:), and the routed system across routing thresholds. 
    Each routed point corresponds to a threshold value, shown directly below the marker. 
    The routed point at $\mathrm{thr}=0.80$ is aligned with the values reported in the table 2. 
    Lower cost and higher score are preferred.
    }
    \label{fig:tradeoff_qwen35_qwen3vl}
\end{figure*}

\section{Training Data and Label Construction}
\label{app:training_data}

\subsection{Capability Head Data}
\label{app:global_data}

The \textbf{Capability Head} is trained on a 120K-example mixture designed to induce a control signal that generalizes across modalities, topics, and task styles. Our goal is not to learn a verifier specialized to a narrow regime such as math reasoning or hallucination detection. Rather, we want a lightweight head that reads the hidden-state trajectory of a frozen model and estimates whether that model is adequate for the current instance under the current setup. For this signal to be useful in practice, it must remain informative across both vision and text, across reasoning and parametric knowledge, and across question answering, grounding, tool use, and more agentic interaction patterns.

To this end, we train on a deliberately mixed dataset spanning visual question answering, science and diagram reasoning, chart and document understanding, screen understanding, UI grounding, multimodal and text-only reasoning, open-domain factual QA, and multi-turn tool-use data. Much of the visual portion is accessed through \textit{FineVision} \citep{wiedmann2025finevision}, which provides a unified interface over heterogeneous multimodal sources. This breadth is intentional. By training on a mixed distribution rather than a single task family, we encourage the head to learn a broader adequacy signal from hidden-state traces rather than a domain-specific notion of confidence. As we show in Appendix~\ref{app:training_mix_ablation}, this breadth is important for transfer: narrowing the training distribution yields a substantially less reliable control signal on a deployment-oriented benchmark.

For each prompt, we query the frozen backbone model, record its generated output and aligned hidden-state trajectory, and score the generation against the task reference using an external LLM-based evaluator. The resulting scalar score in $[0,1]$ is used as the supervision target for the Capability Head.

\begin{figure}[t]
    \centering
    \includegraphics[width=0.92\linewidth]{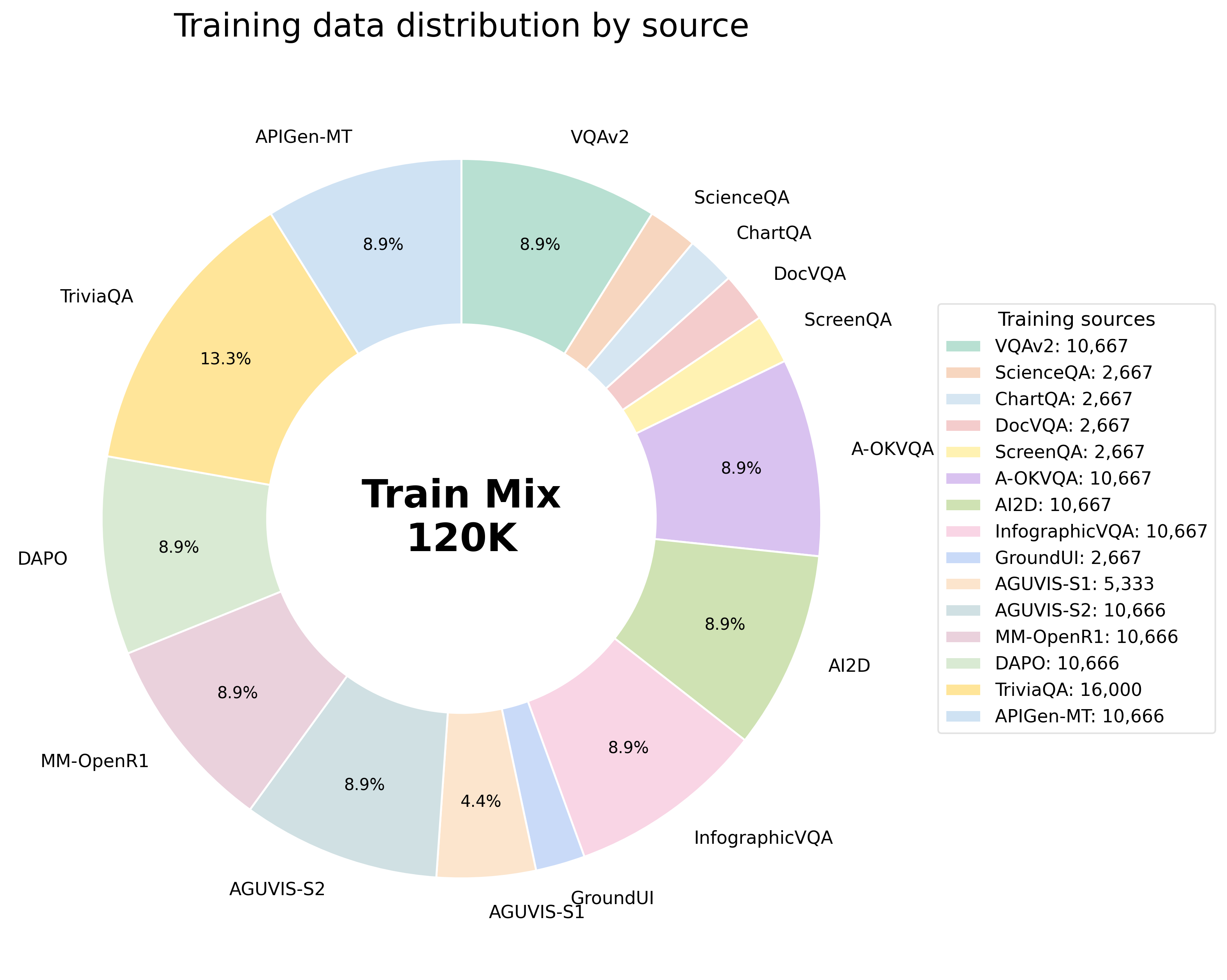}
    \caption{Composition of the 120K training mixture for the Capability Head. The mixture intentionally spans visual QA, reasoning, parametric knowledge, grounding, tool use, and agentic interaction across both vision and text.}
    \label{fig:training_mix}
\end{figure}

\paragraph{Training sources.}
The mixture includes the following datasets.
\begin{itemize}
    \item \textbf{VQAv2} \citep{goyal2017making}: broad visual question answering over natural images, providing general grounded QA examples.
    \item \textbf{ScienceQA} \citep{lu2022learn}: science-oriented multimodal question answering, adding visually grounded scientific and educational reasoning.
    \item \textbf{ChartQA} \citep{masry2022chartqa}: question answering over charts and plots, contributing structured visual reading and lightweight quantitative reasoning.
    \item \textbf{DocVQA} \citep{mathew2021docvqa}: document question answering, contributing OCR-heavy and layout-sensitive visual understanding.
    \item \textbf{ScreenQA} \citep{hsiao2025screenqa}: question answering over screen content, bringing screen-centered visual QA closer to deployment-oriented settings.
    \item \textbf{A-OKVQA} \citep{schwenk2022aokvqa}: knowledge-intensive visual question answering, requiring both image understanding and parametric or external world knowledge.
    \item \textbf{AI2D} \citep{kembhavi2016diagram}: diagram understanding and reasoning, adding a distinct form of structured visual interpretation beyond natural images.
    \item \textbf{InfographicVQA} \citep{mathew2021infographicvqa}: question answering over infographics, combining OCR, layout understanding, and visually embedded factual reasoning.
    \item \textbf{GroundUI} \citep{zheng2024agentstudio}: UI element grounding, contributing fine-grained spatial grounding over interface elements.
    \item \textbf{AGUVIS Stage 1} \citep{xu2024aguvis}: grounded GUI action prediction, introducing action-oriented visual control in interface environments.
    \item \textbf{AGUVIS Stage 2} \citep{xu2024aguvis}: richer GUI interaction data with more agentic structure, extending grounding toward multi-step interface behavior.
    \item \textbf{MM-OpenR1} \citep{evolvinglmmslab2025openr1multimodal}: verified multimodal reasoning data, contributing harder visual reasoning traces.
    \item \textbf{DAPO-Math} \citep{yu2025dapo}: text-only mathematical reasoning, adding long-form reasoning trajectories beyond the visual domain.
    \item \textbf{TriviaQA} \citep{joshi2017triviaqa}: open-domain factual question answering, contributing text-only parametric knowledge and broad topic coverage.
    \item \textbf{APIGen-MT-5k} \citep{prabhakar2025apigenmt}: multi-turn tool-use data, adding tool-calling and agentic task structure beyond standard QA.
\end{itemize}

Figure~\ref{fig:training_mix} summarizes the composition of the training mixture by source.

\subsection{Resolution Head Labels from \textsc{When2Call}}
\label{app:when2call_labels}

\textsc{When2Call} is designed to evaluate tool-calling decision-making, including when to make a tool call, when to ask a follow-up question, and when the query cannot be answered under the current tool setup. However, its released training data is preference-oriented and does not directly provide the exact per-example labels required by our Resolution Head.

We therefore apply an external judge LLM offline to derive a gold resolution decision label for each example in the action space $\mathcal{A}=\{\mathrm{info},\mathrm{tool},\mathrm{cant}\}$. We then prompt the frozen backbone on the same query, record its generated output and hidden-state trajectory, and train the Resolution Head against the derived gold label regardless of whether the backbone's explicit action choice is itself correct. This construction is intentional: it allows the head to recover the correct resolution decision from the model's hidden-state dynamics even when the model's textual behavior is wrong.

\section{Additional Experiments}

\subsection{End-to-End Effect of Prefix-Time Capability Prediction}

Table~\ref{tab:appendix_prefix_routing_tradeoff} extends Table~5 from signal-quality
evaluation to end-to-end routed behavior. While Table~5 measures the quality of the
Capability Control signal itself, here we report the resulting task score and paid API cost
when that signal is used for actual routing. A weaker prefix-time routing signal can lead
to poorer transfer decisions and, in practice, may increase paid API cost by escalating
more cases to the large model. Training directly on partial trajectories improves the
prefix-time adequacy signal and can yield a better early-routing quality--cost tradeoff
than naively applying a full-trajectory-trained head to partial generations.

\begin{table*}[t]
\centering
\small
\setlength{\tabcolsep}{3.0pt}
\renewcommand{\arraystretch}{1.08}
\resizebox{\textwidth}{!}{%
\begin{tabular}{l l l l c c c c c c c}
\toprule
\textbf{Family} & \textbf{System} & \textbf{Train} & \textbf{Infer}
& \textbf{CharXiv} & \textbf{MathVerse} & \textbf{MathVista} & \textbf{ScreenSpot-Pro} & \textbf{SimpleVQA} & \textbf{MMLU-Pro} & \textbf{Overall} \\
\midrule

\multirow{5}{*}{\shortstack[l]{$m_1$: Qwen3-VL-2B-Thk\\$m_2$: Qwen3-VL-32B-Thk}}
& $m_1$: & -- & --
& 0.38 / \$0.00
& 0.74 / \$0.00
& 0.72 / \$0.00
& 0.27 / \$0.00
& 0.33 / \$0.00
& 0.54 / \$0.00
& 0.50 / \$0.00 \\
& $m_2$: & -- & --
& 0.64 / \$1.33
& 0.86 / \$2.75
& 0.84 / \$1.42
& 0.52 / \$6.48
& 0.41 / \$0.68
& 0.72 / \$2.53
& 0.67 / \$15.19 \\
& Routed & Full & Full
& 0.62 / \$1.19 {\scriptsize(\textcolor{red!70!black}{\(\downarrow\)3\%},\,\textcolor{green!60!black}{\(\downarrow\)11\%})}
& 0.85 / \$1.45 {\scriptsize(\textcolor{red!70!black}{\(\downarrow\)1\%},\,\textcolor{green!60!black}{\(\downarrow\)47\%})}
& 0.84 / \$0.96 {\scriptsize(\(\rightarrow\)0\%,\,\textcolor{green!60!black}{\(\downarrow\)32\%})}
& 0.49 / \$4.84 {\scriptsize(\textcolor{red!70!black}{\(\downarrow\)6\%},\,\textcolor{green!60!black}{\(\downarrow\)25\%})}
& 0.39 / \$0.40 {\scriptsize(\textcolor{red!70!black}{\(\downarrow\)5\%},\,\textcolor{green!60!black}{\(\downarrow\)41\%})}
& 0.72 / \$2.22 {\scriptsize(\(\rightarrow\)0\%,\,\textcolor{green!60!black}{\(\downarrow\)12\%})}
& 0.65 / \$11.06 {\scriptsize(\textcolor{red!70!black}{\(\downarrow\)3\%},\,\textcolor{green!60!black}{\(\downarrow\)27\%})} \\
& Routed & Full & Prefix-200
& 0.45 / \$0.30 {\scriptsize(\textcolor{red!70!black}{\(\downarrow\)30\%},\,\textcolor{green!60!black}{\(\downarrow\)77\%})}
& 0.82 / \$0.80 {\scriptsize(\textcolor{red!70!black}{\(\downarrow\)5\%},\,\textcolor{green!60!black}{\(\downarrow\)71\%})}
& 0.79 / \$0.57 {\scriptsize(\textcolor{red!70!black}{\(\downarrow\)6\%},\,\textcolor{green!60!black}{\(\downarrow\)60\%})}
& 0.45 / \$4.31 {\scriptsize(\textcolor{red!70!black}{\(\downarrow\)13\%},\,\textcolor{green!60!black}{\(\downarrow\)34\%})}
& 0.38 / \$0.37 {\scriptsize(\textcolor{red!70!black}{\(\downarrow\)7\%},\,\textcolor{green!60!black}{\(\downarrow\)46\%})}
& 0.68 / \$0.96 {\scriptsize(\textcolor{red!70!black}{\(\downarrow\)6\%},\,\textcolor{green!60!black}{\(\downarrow\)62\%})}
& 0.59 / \$7.31 {\scriptsize(\textcolor{red!70!black}{\(\downarrow\)12\%},\,\textcolor{green!60!black}{\(\downarrow\)52\%})} \\
& Routed & Prefix-200 & Prefix-200
& 0.59 / \$1.09 {\scriptsize(\textcolor{red!70!black}{\(\downarrow\)8\%},\,\textcolor{green!60!black}{\(\downarrow\)18\%})}
& 0.85 / \$1.75 {\scriptsize(\textcolor{red!70!black}{\(\downarrow\)1\%},\,\textcolor{green!60!black}{\(\downarrow\)36\%})}
& 0.83 / \$1.10 {\scriptsize(\textcolor{red!70!black}{\(\downarrow\)1\%},\,\textcolor{green!60!black}{\(\downarrow\)23\%})}
& 0.48 / \$5.26 {\scriptsize(\textcolor{red!70!black}{\(\downarrow\)8\%},\,\textcolor{green!60!black}{\(\downarrow\)19\%})}
& 0.38 / \$0.40 {\scriptsize(\textcolor{red!70!black}{\(\downarrow\)7\%},\,\textcolor{green!60!black}{\(\downarrow\)41\%})}
& 0.77 / \$2.03 {\scriptsize(\textcolor{green!60!black}{\(\uparrow\)7\%},\,\textcolor{green!60!black}{\(\downarrow\)20\%})}
& 0.65 / \$11.63 {\scriptsize(\textcolor{red!70!black}{\(\downarrow\)3\%},\,\textcolor{green!60!black}{\(\downarrow\)23\%})} \\
\midrule

\multirow{5}{*}{\shortstack[l]{$m_1$: Qwen3-VL-4B-Thk\\$m_2$: Qwen3-VL-32B-Thk}}
& $m_1$: & -- & --
& 0.53 / \$0.00
& 0.82 / \$0.00
& 0.78 / \$0.00
& 0.37 / \$0.00
& 0.37 / \$0.00
& 0.65 / \$0.00
& 0.59 / \$0.00 \\
& $m_2$: & -- & --
& 0.64 / \$1.33
& 0.86 / \$2.75
& 0.84 / \$1.42
& 0.52 / \$6.48
& 0.41 / \$0.68
& 0.73 / \$2.47
& 0.67 / \$15.13 \\
& Routed & Full & Full
& 0.63 / \$0.88 {\scriptsize(\textcolor{red!70!black}{\(\downarrow\)2\%},\,\textcolor{green!60!black}{\(\downarrow\)34\%})}
& 0.86 / \$0.98 {\scriptsize(\(\rightarrow\)0\%,\,\textcolor{green!60!black}{\(\downarrow\)64\%})}
& 0.84 / \$0.92 {\scriptsize(\(\rightarrow\)0\%,\,\textcolor{green!60!black}{\(\downarrow\)35\%})}
& 0.49 / \$3.88 {\scriptsize(\textcolor{red!70!black}{\(\downarrow\)6\%},\,\textcolor{green!60!black}{\(\downarrow\)40\%})}
& 0.41 / \$0.31 {\scriptsize(\(\rightarrow\)0\%,\,\textcolor{green!60!black}{\(\downarrow\)54\%})}
& 0.72 / \$1.43 {\scriptsize(\textcolor{red!70!black}{\(\downarrow\)1\%},\,\textcolor{green!60!black}{\(\downarrow\)42\%})}
& 0.66 / \$8.40 {\scriptsize(\textcolor{red!70!black}{\(\downarrow\)1\%},\,\textcolor{green!60!black}{\(\downarrow\)45\%})} \\
& Routed & Full & Prefix-200
& 0.55 / \$0.12 {\scriptsize(\textcolor{red!70!black}{\(\downarrow\)14\%},\,\textcolor{green!60!black}{\(\downarrow\)91\%})}
& 0.85 / \$0.58 {\scriptsize(\textcolor{red!70!black}{\(\downarrow\)1\%},\,\textcolor{green!60!black}{\(\downarrow\)79\%})}
& 0.82 / \$0.84 {\scriptsize(\textcolor{red!70!black}{\(\downarrow\)2\%},\,\textcolor{green!60!black}{\(\downarrow\)41\%})}
& 0.49 / \$4.68 {\scriptsize(\textcolor{red!70!black}{\(\downarrow\)6\%},\,\textcolor{green!60!black}{\(\downarrow\)28\%})}
& 0.41 / \$0.34 {\scriptsize(\(\rightarrow\)0\%,\,\textcolor{green!60!black}{\(\downarrow\)50\%})}
& 0.77 / \$1.31 {\scriptsize(\textcolor{green!60!black}{\(\uparrow\)5\%},\,\textcolor{green!60!black}{\(\downarrow\)47\%})}
& 0.65 / \$7.86 {\scriptsize(\textcolor{red!70!black}{\(\downarrow\)3\%},\,\textcolor{green!60!black}{\(\downarrow\)48\%})} \\
& Routed & Prefix-200 & Prefix-200
& 0.57 / \$0.46 {\scriptsize(\textcolor{red!70!black}{\(\downarrow\)11\%},\,\textcolor{green!60!black}{\(\downarrow\)65\%})}
& 0.84 / \$1.08 {\scriptsize(\textcolor{red!70!black}{\(\downarrow\)2\%},\,\textcolor{green!60!black}{\(\downarrow\)61\%})}
& 0.83 / \$0.93 {\scriptsize(\textcolor{red!70!black}{\(\downarrow\)1\%},\,\textcolor{green!60!black}{\(\downarrow\)35\%})}
& 0.50 / \$5.00 {\scriptsize(\textcolor{red!70!black}{\(\downarrow\)4\%},\,\textcolor{green!60!black}{\(\downarrow\)23\%})}
& 0.40 / \$0.31 {\scriptsize(\textcolor{red!70!black}{\(\downarrow\)2\%},\,\textcolor{green!60!black}{\(\downarrow\)54\%})}
& 0.78 / \$1.60 {\scriptsize(\textcolor{green!60!black}{\(\uparrow\)7\%},\,\textcolor{green!60!black}{\(\downarrow\)35\%})}
& 0.65 / \$9.38 {\scriptsize(\textcolor{red!70!black}{\(\downarrow\)3\%},\,\textcolor{green!60!black}{\(\downarrow\)38\%})} \\
\midrule

\multirow{5}{*}{\shortstack[l]{$m_1$: Gemma-4B-Thk\\$m_2$: Gemma-31B-Thk}}
& $m_1$: & -- & --
& 0.42 / \$0.00
& 0.65 / \$0.00
& 0.67 / \$0.00
& --
& 0.31 / \$0.00
& 0.62 / \$0.00
& 0.53 / \$0.00 \\
& $m_2$: & -- & --
& 0.63 / \$0.62
& 0.87 / \$1.34
& 0.80 / \$0.81
& --
& 0.41 / \$0.39
& 0.78 / \$1.05
& 0.70 / \$4.21 \\
& Routed & Full & Full
& 0.59 / \$0.43 {\scriptsize(\textcolor{red!70!black}{\(\downarrow\)6\%},\,\textcolor{green!60!black}{\(\downarrow\)31\%})}
& 0.84 / \$1.02 {\scriptsize(\textcolor{red!70!black}{\(\downarrow\)3\%},\,\textcolor{green!60!black}{\(\downarrow\)24\%})}
& 0.76 / \$0.44 {\scriptsize(\textcolor{red!70!black}{\(\downarrow\)5\%},\,\textcolor{green!60!black}{\(\downarrow\)46\%})}
& --
& 0.41 / \$0.29 {\scriptsize(\(\rightarrow\)0\%,\,\textcolor{green!60!black}{\(\downarrow\)26\%})}
& 0.76 / \$0.71 {\scriptsize(\textcolor{red!70!black}{\(\downarrow\)3\%},\,\textcolor{green!60!black}{\(\downarrow\)32\%})}
& 0.67 / \$2.89 {\scriptsize(\textcolor{red!70!black}{\(\downarrow\)4\%},\,\textcolor{green!60!black}{\(\downarrow\)31\%})} \\
& Routed & Full & Prefix-200
& 0.50 / \$0.21 {\scriptsize(\textcolor{red!70!black}{\(\downarrow\)21\%},\,\textcolor{green!60!black}{\(\downarrow\)66\%})}
& 0.78 / \$0.77 {\scriptsize(\textcolor{red!70!black}{\(\downarrow\)10\%},\,\textcolor{green!60!black}{\(\downarrow\)43\%})}
& 0.72 / \$0.31 {\scriptsize(\textcolor{red!70!black}{\(\downarrow\)10\%},\,\textcolor{green!60!black}{\(\downarrow\)62\%})}
& --
& 0.38 / \$0.22 {\scriptsize(\textcolor{red!70!black}{\(\downarrow\)7\%},\,\textcolor{green!60!black}{\(\downarrow\)44\%})}
& 0.76 / \$0.27 {\scriptsize(\textcolor{red!70!black}{\(\downarrow\)3\%},\,\textcolor{green!60!black}{\(\downarrow\)74\%})}
& 0.63 / \$1.78 {\scriptsize(\textcolor{red!70!black}{\(\downarrow\)10\%},\,\textcolor{green!60!black}{\(\downarrow\)58\%})} \\
& Routed & Prefix-200 & Prefix-200
& 0.58 / \$0.49 {\scriptsize(\textcolor{red!70!black}{\(\downarrow\)8\%},\,\textcolor{green!60!black}{\(\downarrow\)21\%})}
& 0.84 / \$1.11 {\scriptsize(\textcolor{red!70!black}{\(\downarrow\)3\%},\,\textcolor{green!60!black}{\(\downarrow\)17\%})}
& 0.74 / \$0.39 {\scriptsize(\textcolor{red!70!black}{\(\downarrow\)8\%},\,\textcolor{green!60!black}{\(\downarrow\)52\%})}
& --
& 0.41 / \$0.30 {\scriptsize(\(\rightarrow\)0\%,\,\textcolor{green!60!black}{\(\downarrow\)23\%})}
& 0.82 / \$0.59 {\scriptsize(\textcolor{green!60!black}{\(\uparrow\)5\%},\,\textcolor{green!60!black}{\(\downarrow\)44\%})}
& 0.68 / \$2.88 {\scriptsize(\textcolor{red!70!black}{\(\downarrow\)3\%},\,\textcolor{green!60!black}{\(\downarrow\)32\%})} \\
\bottomrule
\end{tabular}}
\caption{\textbf{End-to-end routed performance under the three prefix-time regimes from Table~\ref{tab:early_global_head}.}
Table~\ref{tab:early_global_head} evaluates the quality of the Capability signal itself; this appendix table shows the corresponding effect on routed task score and paid API cost. \textit{Full / Full} trains and applies the Capability Head on full trajectories, \textit{Full / Prefix-200} applies a full-trajectory-trained head to only the first 200 generated tokens, and \textit{Prefix-200 / Prefix-200} trains and applies the head directly on 200-token prefixes. Each cell reports task score / estimated paid API cost. For routed systems, the inline parenthesis reports the relative change in score and the relative reduction in paid API cost with respect to the corresponding always-large baseline (Model~2).}
\label{tab:appendix_prefix_routing_tradeoff}
\end{table*}

\subsection{Prompt-Level Self-Switching as a Baseline}
\label{app:self_switch_prompt_baseline}

We also compare our latent-control routing against a simple prompt-level self-switching baseline. In this baseline, the smaller model is explicitly instructed not to answer when it is unsure; if no answer is returned, control is transferred to the stronger fallback model. This tests whether the smaller model can trigger useful handoff through its own surface behavior, without an auxiliary latent control signal.

Table~\ref{tab:self_switch_prompt_baseline} reports results for the \textbf{Qwen3.5-9B $\rightarrow$ Qwen3.5-27B-Thk} setting on two representative benchmarks: \textbf{ScreenSpot-Pro} and \textbf{MMLU-Pro}. We compare the prompt-level self-switching baseline against the corresponding single-model and latent-routed systems from the main paper. The self-switching baseline triggers escalation only very rarely---just \textbf{4 / 1581} cases (\textbf{0.25\%}) on ScreenSpot-Pro and \textbf{28 / 1000} cases (\textbf{2.8\%}) on MMLU-Pro. As a result, its performance remains essentially at the small-model level and falls well short of latent-control routing.

This comparison highlights an important distinction. The issue is not whether the smaller model can sometimes express uncertainty in text, but whether it can do so reliably enough to drive useful deployment-time control. In these experiments, relying on prompt-level abstention leads to severe under-escalation, whereas the Capability Head produces a much more effective transfer signal and recovers substantially more of the stronger model's performance.

\begin{table}[t]
\centering
\small
\setlength{\tabcolsep}{7pt}
\renewcommand{\arraystretch}{1.10}
\begin{tabular}{lcc}
\toprule
\multirow{2}{*}{\textbf{System}} & \textbf{ScreenSpot-Pro} & \textbf{MMLU-Pro} \\
& \textit{action acc.} & \textit{accuracy} \\
\midrule
$m_1$: (\textbf{Qwen3.5-9B})                  & 0.47   & 0.71 \\
$m_2$: (\textbf{Qwen3.5-27B-Thk})             & 0.65   & 0.80 \\
Prompt-level self-switching                    & 0.4719 & 0.7030 \\
\rowcolor{blue!8}
Latent routed                                  & 0.64   & 0.78 \\
\bottomrule
\end{tabular}
\caption{\textbf{Prompt-level self-switching versus latent-control routing for Qwen3.5-9B $\rightarrow$ Qwen3.5-27B-Thk.} All entries are benchmark scores. Prompt-level self-switching remains near the small-model baseline on both benchmarks, while latent-control routing recovers substantially more of the stronger model's performance.}
\label{tab:self_switch_prompt_baseline}
\end{table}

\subsection{Model Token Confidence vs.\ Latent Adequacy Signal}
\label{app:confidence_vs_head_screenspot}

\begin{figure*}[t]
    \centering
    \includegraphics[width=\textwidth]{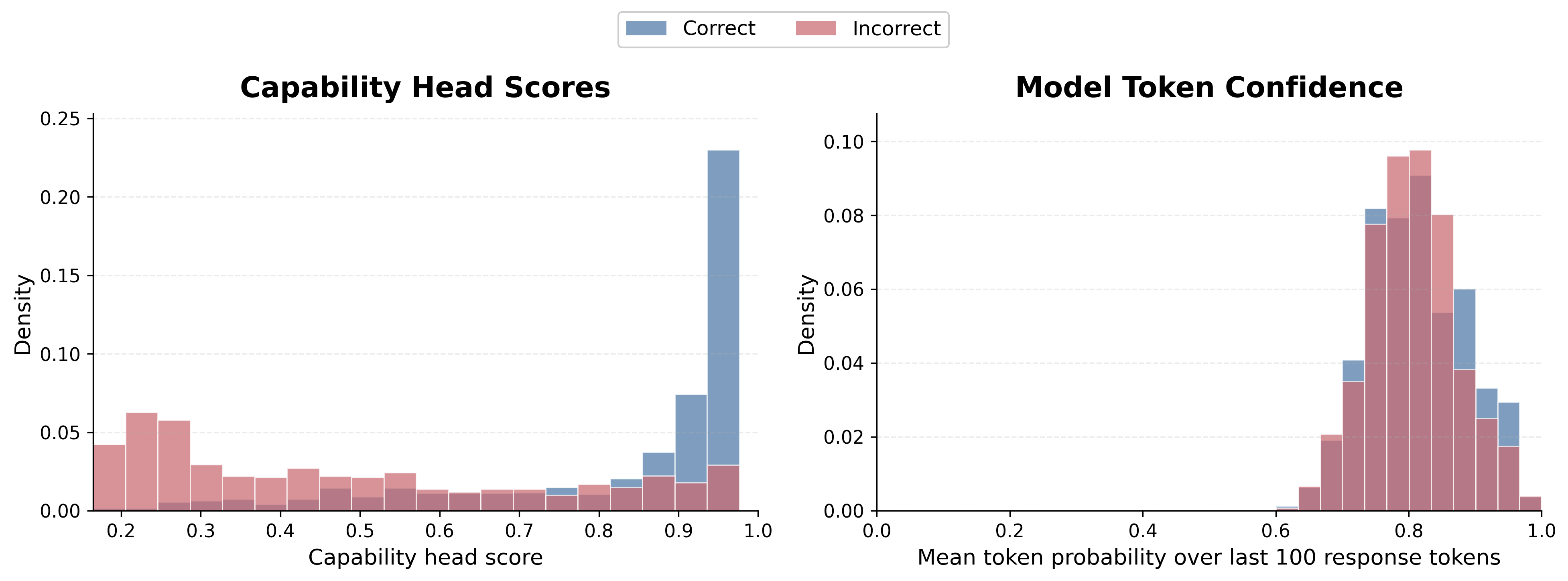}
    \caption{\textbf{Capability Head scores versus model token confidence on Qwen3.5-9B over ScreenSpot-Pro.} The left panel shows the distribution of Capability Head scores for correct and incorrect predictions. The right panel shows the corresponding distribution of the model's mean token probability over the last 100 response tokens. Token confidence exhibits substantial overlap between correct and incorrect cases, making it a weak adequacy signal. In contrast, the Capability Head provides much stronger separation, with correct cases concentrated at high scores and incorrect cases distributed much more broadly.}
    \label{fig:confidence_vs_head_screenspot}
\end{figure*}

We also analyze whether the model's own token-level confidence can serve as a useful substitute for the Capability Head. Figure~\ref{fig:confidence_vs_head_screenspot} compares, on \textbf{Qwen3.5-9B} over \textbf{ScreenSpot-Pro}, the distribution of \textbf{Capability Head} scores against the model's mean token probability over the last 100 response tokens, separating \textbf{correct} and \textbf{incorrect} predictions. The contrast is clear. The model's token confidence shows substantial overlap between correct and incorrect cases, indicating that surface-level confidence is a weak signal for deciding whether the current model is actually adequate for the instance. By contrast, the Capability Head produces much stronger separation: correct cases concentrate at high head scores, while incorrect cases are distributed much more broadly and extend far more heavily into lower-score regions. This suggests that the latent adequacy signal decoded by the head is substantially more informative for deployment-time control than the model's own token confidence. More broadly, the result supports the core premise of the paper: useful control signals need not be reliably expressed in the model's surface probabilities, but can still be recovered from its hidden-state trajectory.

\section{Ablations}
\label{app:ablations}

\subsection{Layer Choice for the Capability Head}
\label{app:global_layer_ablation}

We ablate which hidden-state layer provides the strongest adequacy signal for the Capability Head. Using \textbf{Qwen3-VL-4B-Thk}, we train the same Capability Head on hidden-state traces taken from three representative layers: an early layer (Layer 5), a middle layer, and the final layer. We report the average ROC-AUC, AUPR-C, AUPR-I, and ECE across all evaluation benchmarks. This ablation shows that the final-layer trace provides the strongest overall signal for model adequacy prediction, and motivates the default choice used in the main paper.

\begin{table}[t]
\centering
\small
\setlength{\tabcolsep}{4pt}
\renewcommand{\arraystretch}{1.08}
\begin{tabular}{lcccc}
\toprule
\textbf{Layer Used} & \textbf{ROC-AUC} $\uparrow$ & \textbf{AUPR-C} $\uparrow$ & \textbf{AUPR-I} $\uparrow$ & \textbf{ECE} $\downarrow$ \\
\midrule
Layer 5 & 0.85 & 0.88 & 0.68 & 0.21 \\
Middle  & 0.87 & 0.89 & 0.72 & 0.16 \\
\rowcolor{blue!8}
Final   & 0.86 & 0.91 & 0.75 & 0.14 \\
\bottomrule
\end{tabular}
\caption{Layer-choice ablation for the Capability Head on Qwen3-VL-4B-Thk. We report average ROC-AUC, AUPR-C, AUPR-I, and ECE across all evaluation benchmarks. The final layer is the best overall choice and is therefore used by default.}
\label{tab:global_layer_ablation}
\end{table}

\subsection{Layer Choice for the Resolution Head}
\label{app:local_layer_ablation}

We ablate which hidden-state layer provides the strongest signal for the Resolution Head. Using \textbf{Qwen3-VL-4B} on \textsc{When2Call}, we train the same Resolution Head on traces taken from three representative layers: an early layer (Layer 5), a middle layer, and the final layer. We report F1 and accuracy on the validation split. This ablation shows that a middle-layer trace provides the strongest resolution decision signal, and motivates the default choice used in the main paper.

\begin{table}[t]
\centering
\small
\setlength{\tabcolsep}{6pt}
\renewcommand{\arraystretch}{1.08}
\begin{tabular}{lcc}
\toprule
\textbf{Layer Used} & \textbf{F1} $\uparrow$ & \textbf{Acc} $\uparrow$ \\
\midrule
Layer 5  & 49.1 & 64.2 \\
\rowcolor{blue!8}
Middle   & 52.07 & 70.1 \\
Final    & 50.2 & 66.0 \\
\bottomrule
\end{tabular}
\caption{Layer-choice ablation for the Resolution Head on Qwen3-VL-4B evaluated on \textsc{When2Call}. The middle layer provides the strongest signal for resolution intervention prediction and is therefore used by default.}
\label{tab:local_layer_ablation}
\end{table}

\subsection{Training-Data Breadth}
\label{app:training_mix_ablation}

We next examine whether the breadth of the Capability Head training mixture is genuinely necessary, or whether a useful control signal can be learned from a much narrower domain. This question matters because the goal of the Capability Head is not to predict correctness for one task family, but to estimate \emph{model adequacy} in a way that transfers across various tasks. For that reason, the full 120K training mixture is intentionally constructed to span distinct decision regimes, including visual question answering, grounded perception, reasoning, parametric knowledge, grounding, tool use, and more agentic interaction patterns, rather than serving as a random aggregation of datasets.

To isolate the effect of training-data breadth, we train two Capability Heads on the same frozen \textbf{Qwen3-VL-4B} backbone using the same head architecture and optimization setup. The only difference is the training distribution. One head is trained on the full 120K mixed dataset described in Appendix~A.1. The other is trained only on \textbf{visual-math data}, yielding a substantially narrower signal concentrated on a single reasoning regime.

We evaluate both heads on \textbf{ScreenSpot Pro}, which lies outside that narrow visual-math regime and instead emphasizes visually grounded understanding with grounding- and action-oriented structure. This makes it a useful transfer test: if the Capability Head merely learns a task-specific confidence heuristic tied to visual-math trajectories, transfer should be weak. If, instead, it learns a broader adequacy signal from hidden-state trajectories, then the mixed-data head should generalize substantially better.

Table~\ref{tab:training_mix_ablation} reports results on \textbf{ScreenSpot Pro}. The gap is clear. The head trained only on visual-math data transfers poorly, with substantially weaker discrimination and much worse calibration. In contrast, the head trained on the full mixed dataset remains far more reliable. This result directly supports the logic of the mixture design. Broad training exposure is important not because more data is always better in the abstract, but because the deployment problem itself spans heterogeneous behaviors. Training on a deliberately structured mixture encourages the Capability Head to learn a transferable notion of \emph{when the current model is adequate under the current setup}, rather than a narrow notion of confidence tied to a single task family.

\begin{table}[t]
\centering
\small
\setlength{\tabcolsep}{5pt}
\renewcommand{\arraystretch}{1.08}
\begin{tabular}{lcccc}
\toprule
\multicolumn{5}{c}{\textbf{Evaluation dataset: ScreenSpot Pro}} \\
\midrule
\textbf{Training data for Capability Head} & \textbf{ROC-AUC} $\uparrow$ & \textbf{AUPR-Correct} $\uparrow$ & \textbf{AUPR-Incorrect} $\uparrow$ & \textbf{ECE} $\downarrow$ \\
\midrule
Visual-math only      & 0.62 & 0.58 & 0.61 & 0.59 \\
\rowcolor{blue!8}
Full 120K mixed data  & 0.84 & 0.85 & 0.83 & 0.31 \\
\bottomrule
\end{tabular}
\caption{\textbf{Training-data breadth ablation on ScreenSpot Pro.} Ablation for the Capability Head on the frozen Qwen3-VL-4B backbone. Both heads use the same backbone, latent encoder architecture, and optimization setup; only the training distribution changes. Narrow visual-math-only training transfers poorly to \textbf{ScreenSpot Pro}, while the full mixed-data setting yields a substantially stronger and better-calibrated adequacy signal.}
\label{tab:training_mix_ablation}
\end{table}

\subsection{Prefix-Length Ablation for the Capability Head}
\label{app:prefix_ablation}

We study how the quality of the Capability signal depends on how much of the generated answer is observed. To this end, we train prefix-based variants of the Capability Head on \textbf{Qwen3-VL-4B-Thk}, using only the first $K$ generated tokens, with $K \in \{50, 200, 1000, \text{full}\}$. We evaluate each variant on the same benchmark suite and report the average ROC-AUC, AUPR-C, AUPR-I, and ECE across benchmarks. This ablation measures how early model adequacy becomes reliably recoverable from hidden-state traces.

\begin{table}[t]
\centering
\small
\setlength{\tabcolsep}{4pt}
\renewcommand{\arraystretch}{1.08}
\begin{tabular}{lcccc}
\toprule
\textbf{Prefix Length} & \textbf{ROC-AUC} $\uparrow$ & \textbf{AUPR-C} $\uparrow$ & \textbf{AUPR-I} $\uparrow$ & \textbf{ECE} $\downarrow$ \\
\midrule
50 tokens   & 0.66 & 0.61 & 0.53 & 0.32 \\
200 tokens  & 0.80 & 0.82 & 0.73 & 0.18 \\
1000 tokens & 0.82 & 0.85 & 0.72 & 0.17 \\
Full length & 0.86 & 0.91 & 0.75 & 0.14 \\
\bottomrule
\end{tabular}
\caption{Prefix-length ablation for the Capability Head on Qwen3-VL-4B-Thk. Each row trains and evaluates the head at the same prefix length. We report average ROC-AUC, AUPR-C, AUPR-I, and ECE across benchmarks. Longer prefixes yield stronger adequacy signals and better calibration.}
\label{tab:prefix_ablation_qwen4b}
\end{table}

\section{Cost Estimation Details}
\label{app:cost_estimation}
\vspace{-1em}
We estimate API cost from the observed input and output token counts of each model call. For \textbf{Qwen3-VL} and \textbf{Qwen3.5}, we use the official Alibaba pricing for the corresponding family, matched by model scale and thinking/non-thinking mode. For \textbf{Gemma}, since comparable public API pricing is not available for the models used in our experiments, we use the Qwen3-VL pricing table as a proxy, again matched by scale and inference mode. We then compute cost from the corresponding per-1M-token input and output prices. Since the smaller primary models are run locally in our routed setting, the reported paid API cost reflects only fallback-model usage.

\section{Limitations}
\label{app:limitations}
\vspace{-1em}
The control heads studied here already provide useful deployment-time signals, but their quality still directly affects overall system efficiency. In long-horizon settings, even modest improvements in adequacy prediction or intervention prediction can compound over many steps, leading to better routing decisions, fewer unnecessary escalations, and stronger quality--cost tradeoffs. An important direction for future work is therefore to further improve the quality, robustness, and calibration of these control signals.

\section{Broader View of Latent Control}
\label{app:broader_heads}
\vspace{-1em}
The two heads studied in the main paper should be viewed as concrete instances rather than an exhaustive set. Our broader claim is that hidden-state trajectories provide a reusable substrate for lightweight inference-time control. Prior work has mostly used internal signals for narrow diagnostic goals such as hallucination or correctness estimation. We instead view them as a basis for control.

Once such signals can be read reliably, additional heads can be trained for different deployment needs, user preferences, and application settings, including escalation, clarification, tool use, abstention, confidence shaping, safety filtering, and application-specific routing. In this sense, multi-head latent control is not only a method for the two heads considered here, but also a step toward a broader direction in which frozen models expose lightweight, trainable control interfaces for real-world deployment.

\newpage

\end{document}